\documentclass[lettersize,journal]{IEEEtran}

\usepackage{amsmath,amsfonts}
\usepackage{array}
\usepackage{textcomp}
\usepackage{stfloats}
\usepackage{url}
\usepackage{verbatim}
\usepackage{graphicx}
\usepackage{cleveref}
\crefname{algocf}{alg.}{algs.}
\Crefname{algocf}{Algorithm}{Algorithms}
\usepackage{multirow}
\usepackage{pifont}
\newcommand{\xmark}{\ding{53}}

\usepackage{cite}
\usepackage[ruled,norelsize]{algorithm2e}

\makeatletter
\newcommand{\removelatexerror}{\let\@latex@error\@gobble}
\makeatother
\usepackage{tikz}
\ifCLASSOPTIONcompsoc
    \usepackage[caption=false, font=normalsize, labelfont=sf, textfont=sf]{subfig}
\else
\usepackage[caption=false, font=footnotesize]{subfig}

\usepackage{pgfplots}
\usepackage{pgfplotstable}

\begin{document}
\bibliographystyle{IEEEtran}

\title{A Paradigm Shift in Mouza Map Vectorization: A Human-Machine Collaboration Approach}

\author{Mahir Shahriar Dhrubo$^*$\thanks{\noindent\rule{2cm}{0.4pt}\\
$^*$Both authors contributed equally to this research.
}, Samira Akter$^*$, Anwarul Bashir Shuaib, Md Toki Tahmid, Zahid Hasan, A. B. M. Alim Al Islam}



\maketitle

 \begin{abstract}
 Efficient vectorization of hand-drawn cadastral maps, such as Mouza maps in Bangladesh, poses a significant challenge due to their complex structures. Current manual digitization methods are time-consuming and labor-intensive. Our study proposes a semi-automated approach to streamline the digitization process, saving both time and human resources. Our methodology focuses on separating the plot boundaries and plot identifiers and applying our digitization methodology to convert both of them into vectorized format. To accomplish full vectorization, Convolutional Neural Network (CNN) models are utilized for pre-processing and plot number detection along with our smoothing algorithms based on the diversity of vector maps. The CNN models are trained with our own labeled dataset, generated from the maps, and smoothing algorithms are introduced from the various observations of the map's vector formats. Further human intervention remains essential for precision. We have evaluated our methods on several maps and provided both quantitative and qualitative results with user study. The result demonstrates that our methodology outperforms the existing map digitization processes significantly.
\end{abstract}


\begin{IEEEkeywords}
Mouza Map, Vectorization, Image Processing, Neural Network, GIS\\

\end{IEEEkeywords}

\section{Introduction}
\label{sec:intro}
Maps play a crucial role in a diverse array of applications, which include urban and spatial planning, routing, education, science, and conveying geographic information. \cite{vasile_map_digitization}. In Bangladesh, hand-drawn cadastral maps are used to track the records of land measurement, ownership, plots, and several indicator land areas which are commonly known as mouza maps. Mouza map or Mouza depicts land divisions which serves as an essential tool for land administration and resource management. These hand-drawn maps, containing vital details like land boundaries and land numbers, are pivotal for administrative purposes \cite{nurpost_mouza_map, mouza_wikipedia, sopia_digitization}. Each mouza map has various numbers of plots and each plot has a specific number, essentially an identifier or plot number. Mouza maps provide detailed information about the boundaries of various areas, including their length, area, direction, and other specifics.

Map interpretation, if it is well preserved, clear, and overall in good shape, is a straightforward process that solely relies on the map legend \cite{groom_map_vector}. Visual methods can offer initial insights for preliminary evaluations, however, they are not suitable for in-depth scientific research \cite{map_analysis, map_analysis_2}. Therefore, digitized maps are highly useful for applications such as land management, urban planning, and geographical analysis \cite{sopia_digitization, lithuanian_land, vasile_map_digitization, land_management_initiative}. Traditionally, the digitization of these raster images of maps has been approached through manual vectorization for information retrieval. However, the reliance on manual inspection poses significant challenges. The age and deterioration of these maps further exacerbate the issue, making the process not only time-consuming but also labor-intensive \cite{msdp_digitized_mouza}. This is especially problematic given the vast number of maps and their extensive size.

Over the years, various digitization methods have been applied to maps. However, the digitization of mouza maps has not yet been undertaken. Existing digitization methods primarily focus on raster images, producing vector outputs. Mouza maps, however, contain significant information that must be accurately retrieved, rendering these existing methods unsuitable for mouza maps. Moreover, mouza map vectorization has to be precise, in other words, the vector lines have to be perfectly aligned with raster lines because minor inaccuracies in digitization can lead to substantial real-world discrepancies, a critical factor in land disputes \cite{east_delta_mouza, one_map_policy, land_dispute, land_mafia}. Additionally, most mouza maps are hand-drawn and old, leading to potential information loss during digitization \cite{georeferencing}. Therefore, preserving these details is essential. In other words, mouza map vectorization goes beyond simply converting a raster image into a vector format; it requires exact precision, which, when performed manually can result in several hours or even days of work for a single map.

Currently, the digitization of mouza maps relies on a manual vectorization process. In this process, an operator must zoom in on the raster image to a significant level and add points along the map boundaries. These points must be placed precisely in the center of the boundary lines. Additionally, some sections of the raster maps may appear hazy, requiring the operator to use intuition to accurately position the points. After placing points along all the mouza boundaries, the operator assigns plot numbers based on the original mouza map and then generates a vector file for the specific map. All of this work is typically performed using Geographical Information System, in short, GIS software, such as ArcGIS. It would be significantly more efficient if the entire map could be vectorized with a single click. 

With this objective in mind, in this paper, we propose a novel methodology to digitize the mouza maps in an automated way which significantly accelerates this digitization process and replaces the manual work of the operator with our proposed automation steps. To the best of our knowledge, the automation of full digitization of Mouza maps has not been previously attempted. The initial step in georeferencing old maps involves scanning the physical paper maps \cite{ijgi8100455, Krejci_georeferencing, Statuto_Cillis_Picuno_2016, stauble_switz, inproceedings, ica-proc-4-38-2021}. Our process begins with scanned versions of these maps. Our methodology has four essential steps (\cref{fig:method_flowchart}). The first step involves a manual screening process and reconstructing the parts of the maps that are hazy or missing. This step also includes separating plot boundaries and digits. In the second step, we generate digits and plot numbers. The third step focuses on vectorizing the boundaries using our refinement algorithms. Finally, in the fourth step, the plot numbers are merged with the vectorized boundaries to create a vectorized file of the mouza maps. However, given the age of the maps and hand-drawn by different individuals, automated processes are not foolproof \cite{forest_mouza}. Hence, manual inspection remains necessary after our process to align the vectorized maps with their raster counterparts accurately and to complete any sections that the automated process may have missed.

We compared and evaluated our approach against the manual vectorization from scratch through user study. Our approach demonstrated time savings of up to $50\%$. For instance, a manual process requiring 25 hours could be accomplished within 13 hours using our approach. This scalability and efficiency are crucial for countries like Bangladesh, where vast numbers of cadastral maps require digitization. While our approach needs human attention, correcting the vector output of our approach takes significantly less time than starting from scratch. To summarize our key contributions are :
\begin{itemize}
    \item Developing a semi-automated way to vectorize the mouza map, where the entire pipeline runs automatically except for pre-processing.
    \item Developing a robust and essential pre-processing stage to fill gaps in the mouza boundaries. This stage is critical because 
    empty regions in the borderlines will interrupt the automatic vectorization process.
    \item Developing a novel methodology for Mouza map vectorization. We have used Optical Character Recognition (OCR) for the first time in literature to digitize cadastral maps.
    \item Introducing smoothing algorithms specially designed for vectorized mouza maps. These algorithms reduce the irregularities, resulting in a more precise vector representation. We have discussed our significant improvements in the experiment and comparison sections.

\end{itemize}
\noindent
\section{Related Work}

Traditionally in cartography practice, topographical and cadastral maps are hand-drawn on paper \cite{yang_contour_lines}. To increase scalability and efficiency, computer systems have become an essential tool for resource management and urban planning with geographical information \cite{LEE2000165}, and the need to digitize analog or hand-drawn maps has increased substantially due to the recent proliferation of GIS technologies \cite{taie2011new}. Over the years, several approaches have been proposed to convert hand-drawn maps into digitized form. Illert et al., \cite{illert1991automatic} describes a software system, automatically digitizing large-scale maps. Studies have been found to perform vectorization of maps without digits or plot numbers, with segmentation \cite{KIM20141262} or with GIS software \cite{digitization_arcGis, Iosifescu, xu2017vectorization}. This system can convert raster data into structured vector data \cite{eken2015automated}. Chen et al., \cite{chen2014development}  have discussed a system developed for automatically converting hand-drawn maps to tactile maps. The process begins with a hand-drawn map, which is then digitized using a scanner or digital camera, and finally translated into digital files (SVG and Edel documents) for generating tactile maps. In both of these approaches, the process of converting raster data into structured vector data might not capture all the nuances of the original maps, leading to potential inaccuracies. Vasile et al., \cite{vasile_map_digitization} proposed descriptive processes for map digitization, no automation was used. Rahim et al., \cite{sopia_digitization} have discussed the legal approaches for land digitization. Piskinaite et al., \cite{lithuanian_land} described the digitization of historical maps, suggesting automation of the digitization process. However, their study lacks a clearly defined structure to implement this digitization process fully. Ahmed et al., \cite{ahmed2024digitizing}, also proposed a digitization method by extracting the cadastral boundary with unsupervised classification algorithms, however, the vectorization process remains manual.

Several works have been done in the area of map vectorization by implementing image processing and deep learning ideas. Bhunia et al., \cite{bhunia2021vectorization} proposed a self-supervised learning technique for sketches and handwriting data, incorporating both rasterized images and vector images. It proposes a self-supervised learning method with two cross-modal translations: Vectorization and Rasterization.  A Layer-wise Image Vectorization (LIVE) method is proposed in the work of Zhou et al., \cite {Ma_2022_CVPR} to convert raster images to SVGs while maintaining image topology. It proposes a model-free approach that avoids needing specific domain knowledge, like fonts and emojis and does not require extensive SVG dataset collection.

Land estimation from an image can be achieved through zooming and the Canny edge method \cite{canny_edge_method}. Akter et al., \cite{akter} have proposed a method for measuring land area, which involves the design of a system employing various image processing techniques. Digital maps find applications in various fields, such as geographical information systems (GIS), satellite imagery analysis, and geology, among others. In a study by Samet et al., \cite{samet} a technique for precise contour lines extraction from scanned geographical maps was demonstrated. Vectorization refers to the process of converting raster images into vector representations. Egiazarian et al.,\cite{rassian} in their work, present an automated approach to reconstruct vector representations of technical drawings. These drawings encompass a variety of elements, including line segments and circular arcs. Chen et al., \cite{vectorization_of_historical_map} propose an automated solution for transforming raster maps into vector objects, with a specific focus on vector objects alone. In contrast, Iosifescu et al., \cite{Iosifescu} present an automatic method aimed at tracking spatial changes in historical maps over time through vectorization of raster maps. Similarly, JiaZhou et al., \cite{jiaZhou} introduce a vectorization method centered on junction analysis, which also includes enhancements for tracking and vectorizing rough lines. Although all these works focus on the vectorization of hand-drawn maps, a lack of concentration focusing particularly on low-resource areas, for example, rural map drawing is noticed.

\section{Research Context}
\label{research_context}\vspace{0.001\linewidth}
\begin{figure}
  \centering
  \begin{tikzpicture}
    \node[anchor=south west,inner sep=0] (image) at (0,0) {\includegraphics[width=8.2 cm, height = 8cm, angle=90]{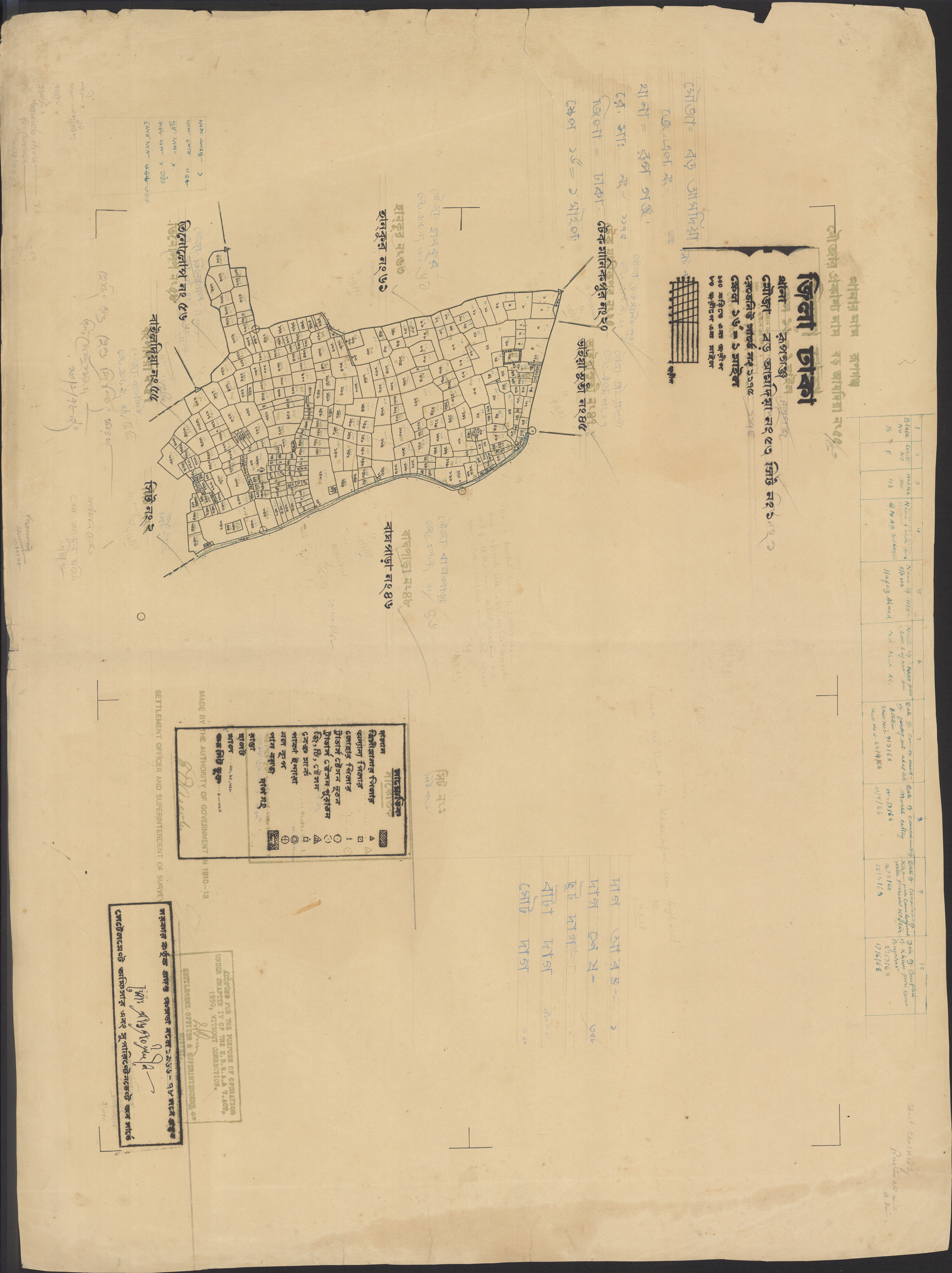}};
    \draw[red] (2.7,3.05) rectangle (2.82,3.2);
     \node[anchor=south west, inner sep=0] at (4,4) { \includegraphics[width=1cm, height=1cm]{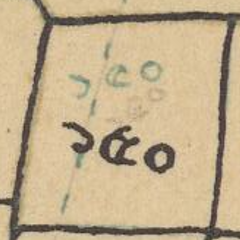}};
    \draw[->,red] (2.82,3.15) -- (4,4);
     \draw[->,red] (4.5,4.40) -- (5.3,4.8);
     
    \draw[red] (1.6,5.7) rectangle (3,7.2);
    \draw[red] (4,4) rectangle (5,5);
     \draw[red] (4.5,1.5) rectangle (5.4,3.5) ;
     \draw[red] (5.6,0.9) rectangle (7.3,1.6) ;
     
    \node[red, align=left] at (4.98,6.5) {
    Map details- (location)\\
    Scale: 16 inches = 1 mile};

    \node[red]  at (5.5,4.4){Plot};
    \node[red]  at (6.5,5){Plot Number: 150};
    \node[red]  at (6.2,2.5){Legend};
    \node[red]  at (6.4,0.6){Authority signature};
  \end{tikzpicture}
  \caption{Image of a mouza map, containing 300 plots}
  \vspace{-10pt}
  \label{fig:mouza_map}
\end{figure}
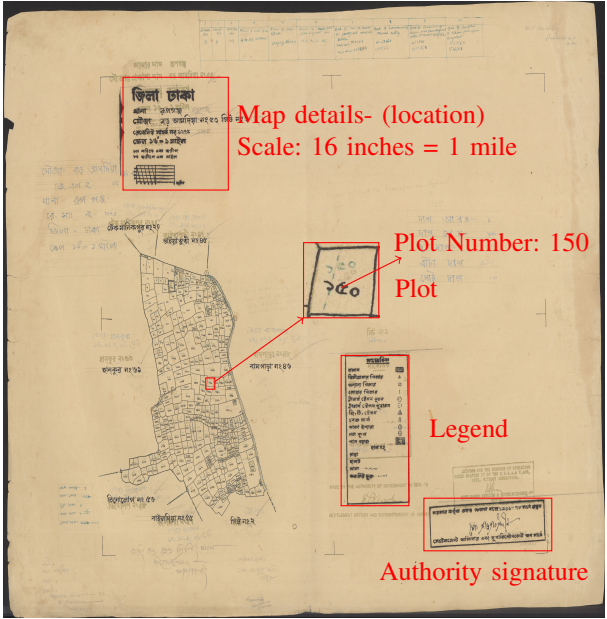
A full mouza map is illustrated in \cref{fig:mouza_map}, featuring a singular plot with an embedded Mouza number or plot number. This map encompasses over 300 such plots with a unique plot number, alongside additional map details, legend, and authority signature. The map details section provides information about the map's location and specifies the scale, which in this case is 16 inches to 1 mile in the illustrated map. The scale implies that even slight errors in the vector map can lead to considerable deviations. The legend section of the map includes various symbols and the plot numbers which are present in the map. The plot numbers can range from less than 100 to 4000 and, the plot areas are variable, potentially resulting in larger maps. To accurately vectorize this map (\cref{fig:mouza_map}), it requires $7$ to $9$ hours depending on the individual's proficiency. The sizeable time requirement for a single map serves the perspective on vectorizing larger maps with 2000 or more plots, discussed in the experiment result (\cref{subsec:comparison}). This underscores our motivation to minimize the effort of a human operator as well as resources to vectorize mouza maps given their substantial volume, numbering in the tens of thousands.
\SetKwComment{Comment}{/* }{ */}
\section{Methodology}
\label{sec:methodology}

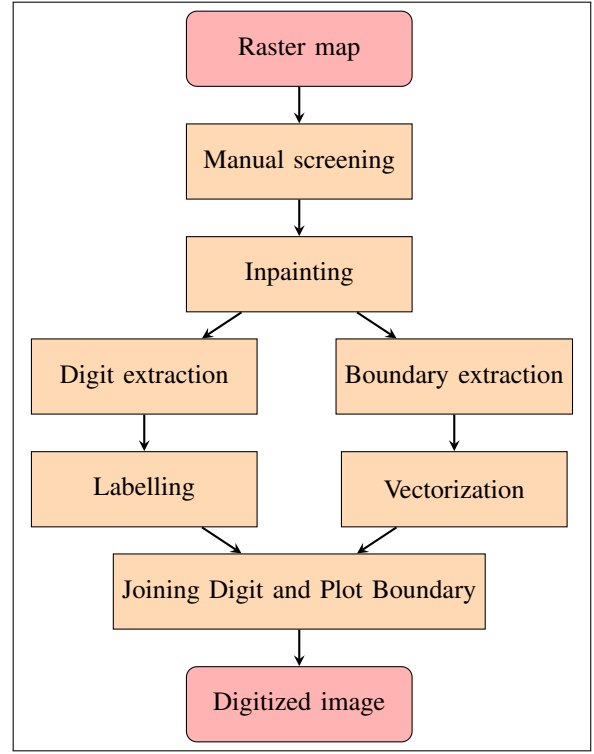
\begin{figure}
    \centering
    \scalebox{1}{\fbox{
    \begin{tikzpicture}[node distance=1.5cm]
    \tikzstyle{startstop} = [rectangle, rounded corners, minimum width=3cm, minimum height=1cm, text centered, draw=black, fill=red!30]
    \tikzstyle{process} = [rectangle, minimum width=3cm, minimum height=1cm, text centered, draw=black, fill=orange!30]
    \tikzstyle{decision} = [diamond, minimum width=3cm, minimum height=1cm, text centered, draw=black, fill=green!30]
    \tikzstyle{arrow} = [thick,->,>=stealth]

    \node (rastermap) [startstop] {Raster map};
\node (preprocess) [process, below of=rastermap] {Manual screening};
\node (linejoin) [process, below of=preprocess] {Inpainting};
\node (digits) [process, below left of=linejoin, xshift=-1cm, yshift=-0.3cm] {Digit extraction};
\node (boundary) [process, below right of=linejoin, xshift=1cm, yshift=-0.3cm] {Boundary extraction};
\node (labeler) [process, below of=digits] {Labelling};
\node (vectorize) [process, below of=boundary] {Vectorization};
\node (joinparse) [process, below left of=vectorize, xshift=-1cm, yshift=-0.3cm] {Joining Digit and Plot Boundary};
\node (digitized) [startstop, below of=joinparse] {Digitized image};

\draw [arrow] (rastermap) -- (preprocess);
\draw [arrow] (preprocess) -- (linejoin);
\draw [arrow] (linejoin) -- (digits);
\draw [arrow] (linejoin) -- (boundary);
\draw [arrow] (digits) -- (labeler);
\draw [arrow] (boundary) -- (vectorize);
\draw [arrow] (labeler) -- (joinparse);
\draw [arrow] (vectorize) -- (joinparse);
\draw [arrow] (joinparse) -- (digitized);
    \end{tikzpicture}
    }}
    \caption{Step-by-step outline of our proposed methodology}
    \vspace{-10pt}
    \label{fig:method_flowchart}
\end{figure}

In this following section, we discuss our proposed pipeline for converting a raster image into a vectorized map. The workflow, detailed in \cref{fig:method_flowchart}, involves manual refinement, gap filling using a line inpainting model, and separation of the map into boundary lines and numerical digits. We then aggregate plot numbers after identifying numerical digits and vectorize the boundary lines. A refinement step is applied to the vectorized output using our smoothing algorithms. Subsequently, we polygonize the refined vector map and join the identified plot numbers with their corresponding polygon, which represents a plot in mouza maps.

\subsection{Pre-processing}
\label{preprocessing}
Mouza maps contain various symbols, including markers for locations and legend, which are irrelevant to our digitization objectives. We conduct an initial manual screening where we retain only the crucial elements: the land boundaries and plot numbers using standard image editing tools. This manual screening is not exhaustive.\\
\subsubsection{Boundary Inpainting}
\label{susubsec:boundary_inpainting}
\begin{figure}
    \centering
    \includegraphics[width=\linewidth]{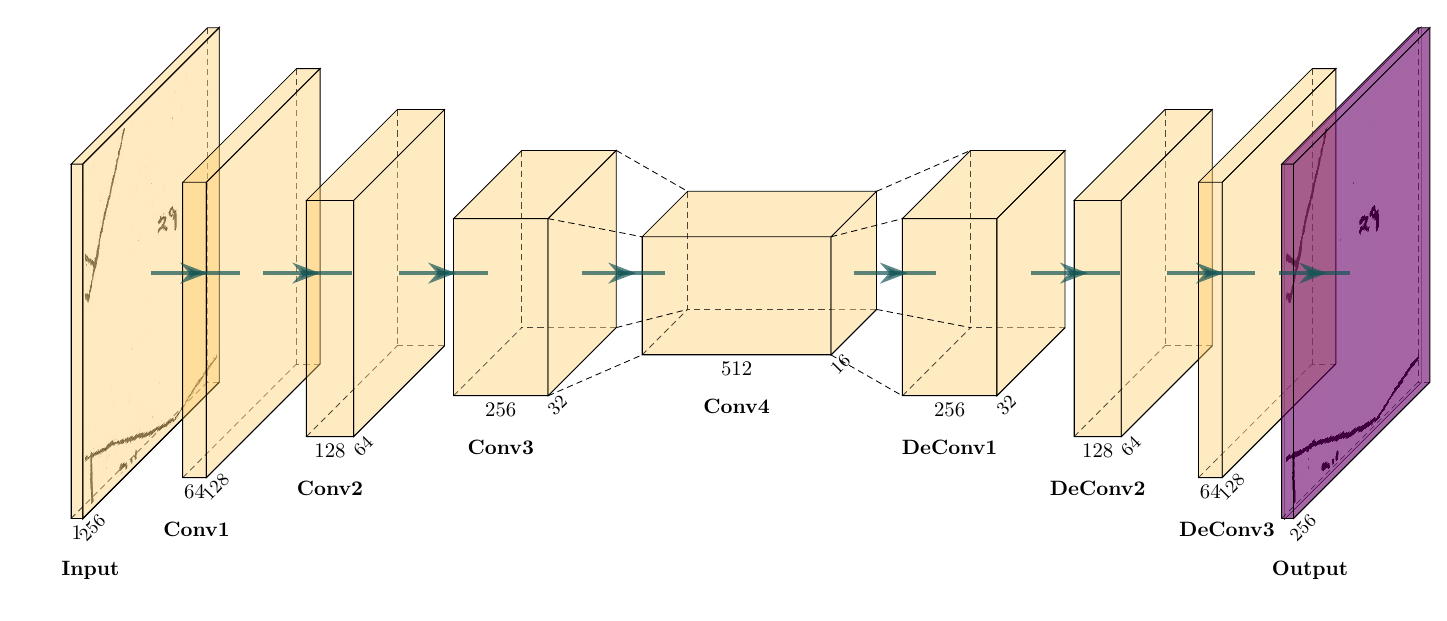}
    \caption{Boundary inpainting model architecture}
    \label{fig:architecture_inpaint}
\end{figure}
\vspace{-10pt}
In several maps, the boundary lines of plots are incomplete or unclear. It is essential to fill in the missing regions for our digits and plot separation. To address this, we have developed a context-aware deep learning model. Our inpainting approach is similar to the works of Sasaki et. al., \cite{Sasaki_2017_CVPR}. This approach utilizes full convolutional layers, that were introduced for semantic segmentation tasks \cite{Noh_2015_ICCV, Long_2015_CVPR}.
Given the substantial size of the original JPG maps, we divide them into overlapping sections of $300\times300$ pixels and process these through our model.

The architecture of our model consists of four convolutional layers, alongside four deconvolutional layers, illustrated in \cref{fig:architecture_inpaint}. The convolutional layers have respectively 64, 128, 256, and 512 filters and the doconvolutional layers have 256, 128, 64, and 1 filters respectively. A kernel size of $4 \times 4$ along with 1-pixel padding and a stride of 2-pixel is used in all the convolution layers to reduce to half size. This model is designed to analyze the lines and fill in any gaps or hazy parts of the plot boundaries. Following the inpainting process, we reassemble the segments to produce a restored cadastral map.

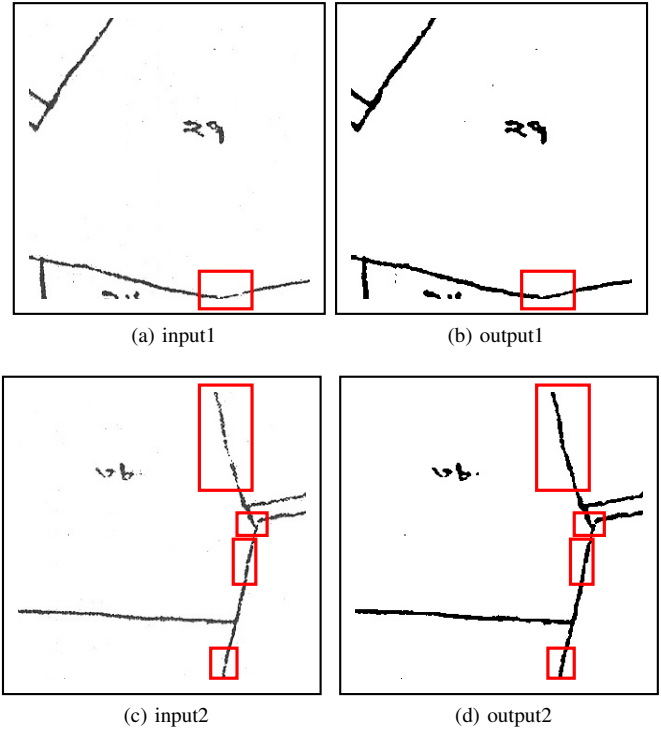
\begin{figure}
    \centering

 \subfloat[input1]{%
  \begin{tikzpicture}
    \node[inner sep=0pt] (image) at (0,0) {\includegraphics[width=0.42\linewidth]{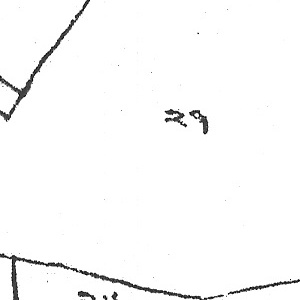}};
    \def\padding{0.2cm}
    \draw[black, thick] 
      ([xshift=-\padding, yshift=-\padding]image.south west) 
        rectangle 
      ([xshift=\padding, yshift=\padding]image.north east);
      \draw[red, very thick] (0.4,-1.5)rectangle (1.1,-2);
    
  \end{tikzpicture}
    }
  \subfloat[output1]{%
         \begin{tikzpicture}
        \node[inner sep=0pt] (image) at (0,0) {\includegraphics[width=0.42\linewidth]{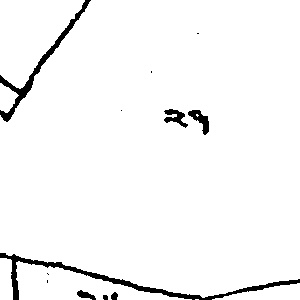}};
        \def\padding{0.2cm}
        \draw[black, thick] 
          ([xshift=-\padding, yshift=-\padding]image.south west) 
            rectangle 
          ([xshift=\padding, yshift=\padding]image.north east);
         \draw[red, very thick] (0.4,-1.5)rectangle (1.1,-2);
      \end{tikzpicture}
     }
    \\
  \subfloat[input2]{%
        \begin{tikzpicture}
        \node[inner sep=0pt] (image) at (0,0) {\includegraphics[width=0.43\linewidth]{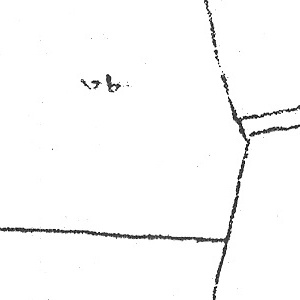}};
        \def\padding{0.2cm}
        \draw[black, thick] 
          ([xshift=-\padding, yshift=-\padding]image.south west) 
            rectangle 
          ([xshift=\padding, yshift=\padding]image.north east);
        \draw[red, very thick] (0.5,0.6)rectangle (1.2,2);
        \draw[red, very thick] (1,0)rectangle (1.4,0.3);
        \draw[red, very thick] (0.95,-0.05)rectangle (1.25,-0.65);
        \draw[red, very thick] (0.65,-1.5)rectangle (1,-1.9);
         \end{tikzpicture}
        }
    \hfill
  \subfloat[output2]{%
        \begin{tikzpicture}
        \node[inner sep=0pt] (image) at (0,0) {\includegraphics[width=0.43\linewidth]{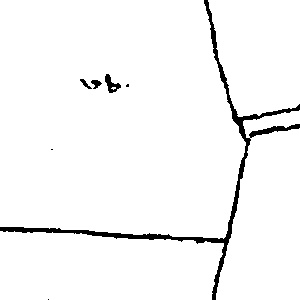}};
        \def\padding{0.2cm}
        \draw[black, thick] 
          ([xshift=-\padding, yshift=-\padding]image.south west) 
            rectangle 
          ([xshift=\padding, yshift=\padding]image.north east);
        \draw[red, very thick] (0.5,0.6)rectangle (1.2,2);
         \draw[red, very thick] (1,0)rectangle (1.4,0.3);
         \draw[red, very thick] (0.95,-0.05)rectangle (1.25,-0.65);
         \draw[red, very thick] (0.65,-1.5)rectangle (1,-1.9);
         \end{tikzpicture}
  
        }
  \caption{Boundary inpainting sample inputs and outputs. Output images are connected i.e., filled the gaps from input}
  \label{fig:gap_1} 
\end{figure}

For training, we prepare a dataset from a variety of cadastral maps, dividing the map images into $300\times300$ patches. The input images are generated by intentionally introducing gaps on the boundary lines whereas the corresponding output images are the original, non-gapped versions, which serve as the ground truth. Our model, trained on a collection of 1,723 such binary images, demonstrated commendable performance for the reconstructions of the small missing or hazy regions of the raster images. Sample results of the reconstruction are showcased in \cref{fig:gap_1}.

\subsubsection{Connected Component Analysis and Separation of Plot Numbers}
\label{subsec:connected-component}

In a raster mouza map, there are several connected components. A connected component is a group of pixels with a value of 1(or 255) in a binary image that is connected by their adjacency or proximity. We observe that the largest one is always the boundary line. We employed connected component analysis to separate the components. Hence, we select the component with the largest area, which represents the boundary lines of the plots. Apart from the largest component, several smaller components contain mostly digits and some noises. Through experimentation, we determined a threshold value that represents the typical area of a single digit. We then filter out the other components based on this threshold to isolate the digits. Each digit was saved along with its corresponding position, i.e., its pixel coordinates, which will later be used for generating the plot number. However, there is one big concern that digits and plot boundaries sometimes overlap in the raster image. In consequence, the boundary component also contains some digits after separation. We manually disconnect the overlapped digits by erasing connecting pixels from the overlapped region, which can be done by any standard image editing tool. Then we repeat the connected component analysis to separate the remaining digits, which we just edited from the boundary.

To summarize, we utilize connected component analysis to identify and separate the largest component, which represents the plot boundary, and using a threshold value we filter out the digits. After manually editing to disconnect any overlapping digits, we repeat the connected component analysis to separate any remaining digits that were not isolated in the first iteration. At this stage, our step-by-step methodology diverges to handle two tasks: the OCR processing of plot numbers and the vectorization of land boundaries.

\subsection{Digit Recognition}
\label{subsec:digit_recognition}
\begin{figure}
        \centering
        \includegraphics[width = \linewidth]{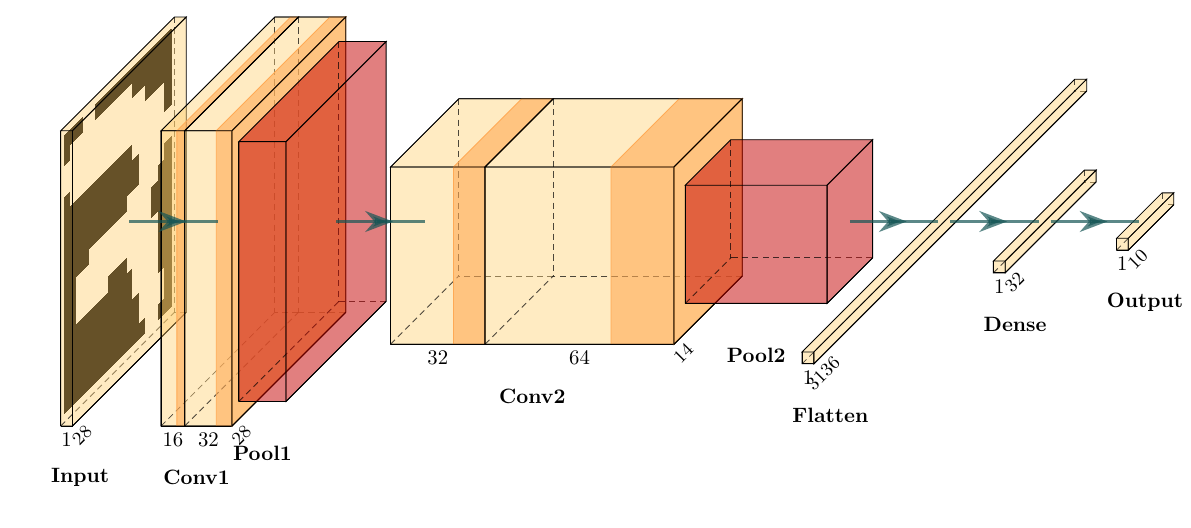}
        \caption{Deep learning model architecture for digit recognition}
    \label{fig:architecture}
\end{figure}
\begin{figure} 
    \centering
  \subfloat[Digit 3]{%
       \includegraphics[width=0.3\linewidth]{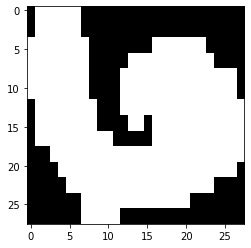}}
    \hfill
  \subfloat[Digit 5]{%
        \includegraphics[width=0.3\linewidth]{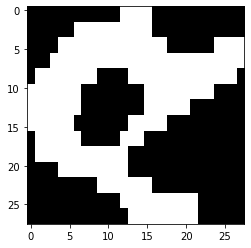}}
    \hfill
  \subfloat[Digit 9]{%
        \includegraphics[width=0.3\linewidth]{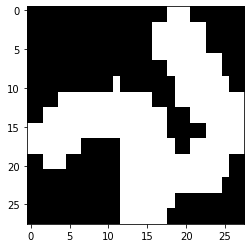}}
  \caption{Samples of digits from the mouza map. Digits 3,5,9 from left to right}
  \label{fig:digits} 
\end{figure}

\cref{subsec:connected-component} allows us to extract and separate the digits within the mouza maps along with their coordinate. Some sample digits are shown in \cref{fig:digits}
Now, to identify the digits, we employ a neural network consisting of four convolutional layers. The model architecture of the digit recognizer is shown in \cref{fig:architecture}. The input to this network is 28$\times$28 digit images from the mouza map. We have prepared a dataset with 229061 digit images for training the model and 4401 images are used for validation. After 100 epochs of training, the digit recognizer provides 95.86\% accuracy. It is to be noted that, training the model with the extracted digits from the mouza map is essential because of the distinct nature of the writing style while annotating a mouza map. 

After identifying each digit, we map the identified labels (0 to 9) to their corresponding position i.e., pixel coordinate. This process yields a collection of digits along with their exact positions. These digit-position pairs will then be integrated with the vectorized plot boundary of the mouza map. 

\subsection{Vectorization}
\label{subsec:vectorization}
\begin{figure}
    \subfloat[]{
        \fbox{\includegraphics[width = 0.45\linewidth]{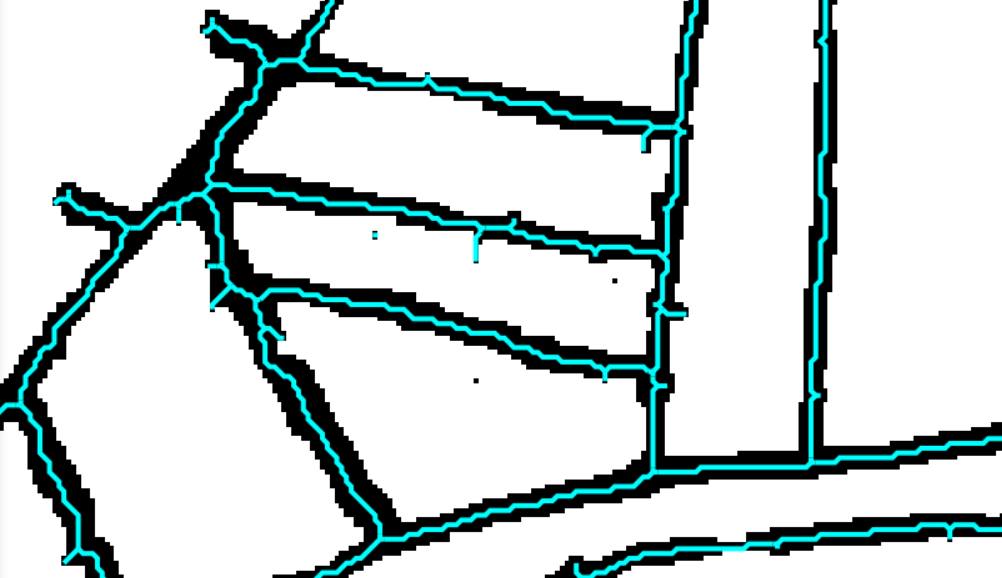}}}
    \hfill
    \subfloat[]{
        \fbox{\includegraphics[width = 0.45\linewidth]{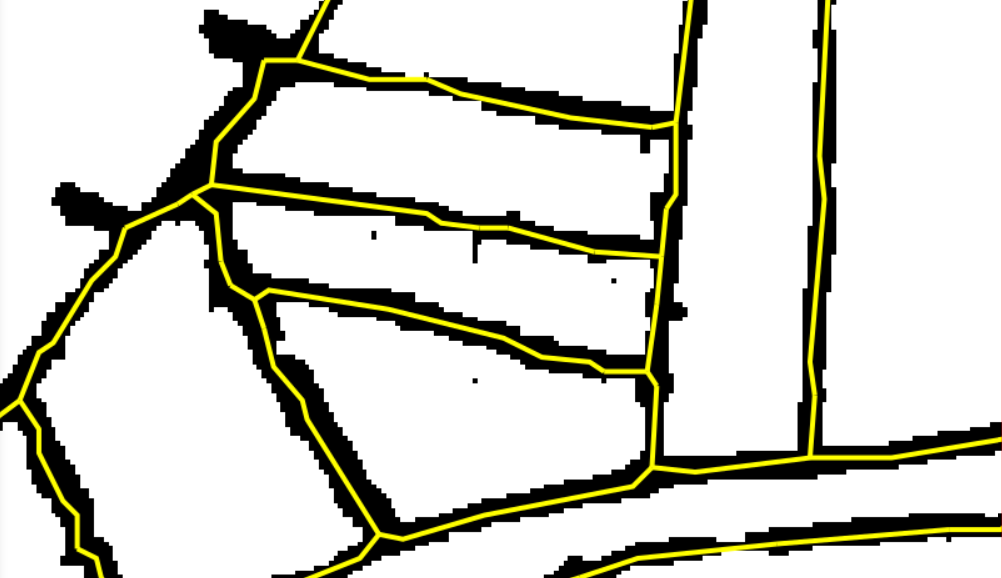}}}
    \caption{Grass GIS outputs. a) The basic output with lots of floating and un-smooth lines b) Floating lines are removed, lines are still un-smooth}
    \label{fig:gis_output}
\end{figure}
We utilize GRASS GIS, controlled through a Python script, to automate this phase \cite{GRASS_GIS_software}. The workflow begins with the importation of the map using \textit{r.in.gdal} \cite{r_in_gdal_grass}, followed by the application of \textit{r.thin} \cite{r_thin_grass} to distill the land boundaries to a single-pixel-width framework. The transition from raster to vector format is achieved using \textit{r.to.vect} \cite{r_to_vect_grass}, which is then followed by a sequence of cleaning operations with \textit{v.clean} \cite{v_clean_grass}.These operations include snapping vertices, excising dangling lines, and simplifying the vector paths by removing superfluous vertices and straightening angles at nodes without altering the vector topology, as shown in \cref{fig:gis_output}. Despite the cleaning process, the vector lines are not smooth and exhibit noticeable irregularities and a staircase effect. We then export the vector file in ASCII text format, which contains the coordinates of the points on the vector map. This format allows us to edit the coordinates and refine the vector lines to achieve a smoother representation. This process requires no human intervention. Thinning, vectorization, and exporting are fully automated using a GIS Python plugin. Following this, our smoothing algorithms are applied to the exported ASCII file.
\subsubsection{Editing ASCII file and Smoothing}
\label{subsubsec:smoothing}
\begin{figure}
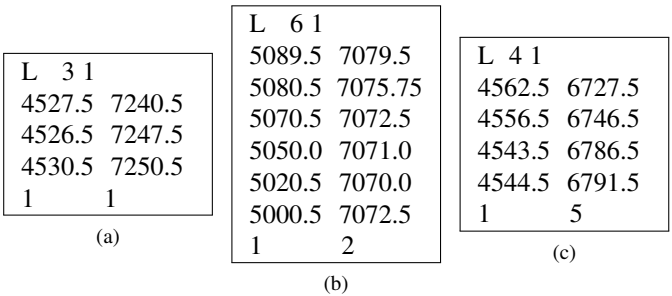

\centering
    \subfloat[]{
         \fbox{
            \begin{minipage}{0.26\linewidth}
        L \hspace{0.2cm}3\hspace{0.1cm}1\\
        4527.5 \hspace{0.1cm}7240.5\\
        4526.5 \hspace{0.1cm}7247.5\\
        4530.5 \hspace{0.1cm}7250.5\\
        1 \hspace{0.8cm}1
        \end{minipage}
        }
    }
    \subfloat[]{
     \fbox{
            \begin{minipage}{0.26\linewidth}
        L \hspace{0.2cm}6\hspace{0.1cm}1\\
        5089.5 \hspace{0.1cm}7079.5\\
        5080.5 \hspace{0.1cm}7075.75\\
        5070.5 \hspace{0.1cm}7072.5\\
        5050.0 \hspace{0.1cm}7071.0\\
        5020.5 \hspace{0.1cm}7070.0\\
        5000.5 \hspace{0.1cm}7072.5\\
        1 \hspace{0.8cm}   2
        \end{minipage}
        }
    }
    \subfloat[]{
       \fbox{
            \begin{minipage}{0.26\linewidth}
         L\hspace{0.2cm}4\hspace{0.1cm}1\\
        4562.5 \hspace{0.1cm}6727.5\\
        4556.5 \hspace{0.1cm}6746.5\\
        4543.5 \hspace{0.1cm}6786.5\\
        4544.5 \hspace{0.1cm}6791.5\\
        1 \hspace{0.8cm}   5
        \end{minipage}
        }
    }
    \caption{Different images from ASCII file each representing a category. These categories have 3, 6, and 4 points respectively}
    \label{fig:ascii}
    \vspace{-10pt}
\end{figure}
The ASCII file holds the coordinates of points constituting the vector map. This file follows a structured format where points are organized into categories. Each category corresponds to a distinct segment of the vector boundary and encompasses multiple points. In \cref{fig:ascii}, various categories are shown, each representing a vector line segment. To form a vector line segment, the points within each category are connected sequentially from the first to the last point. The first and last points are connected to two different intersections. There are no other intersections within the same category, meaning that each category represents a series of points lying between two intersections.

\SetKwComment{Comment}{/* }{ */}
\begin{algorithm}
\caption{Solve join error}\label{alg:correct_join}
\KwData{C (categories)}
\KwResult{$C_n$ (categories after removing join errors)}

$D \gets \{\}$ \Comment*[r]{empty dictionary}
\For{$cat \in C$}{
    $D[cat.leftPoint].append(cat)$\;
    $D[cat.rightPoint].append(cat)$\;
}
\For{$cat \in C$}{
    $lCats \gets D[cat.leftPoint]$\;
    $rCats \gets D[cat.rightPoint]$\;
    \If{$len(lCats) \geq 3$ and $len(rCats) \geq 3$)}{
        $mp \gets getMidPoint(cat.leftPoint, cat.rightPoint)$\;
        \For{$lCat \in lCats$}{
            $lCat.replace(lp, mp)$\;
        }
        \Comment*[r]{repeat loop for \textbf{rCats}}
        $cat.remove \gets true$\;
    }
}
$C_n \gets []$\;
\For{$cat \in C$}{
    \If{$not cat.removed$}{
    $C_n.append(cat)$\;
    }
}
\Return $C_n$\;

\end{algorithm}

\begin{figure}
    \centering
    \subfloat[Four lines join error(left) and solve(right)]{
        \fbox{\includegraphics[width = 0.45\linewidth, height = 0.5\linewidth]{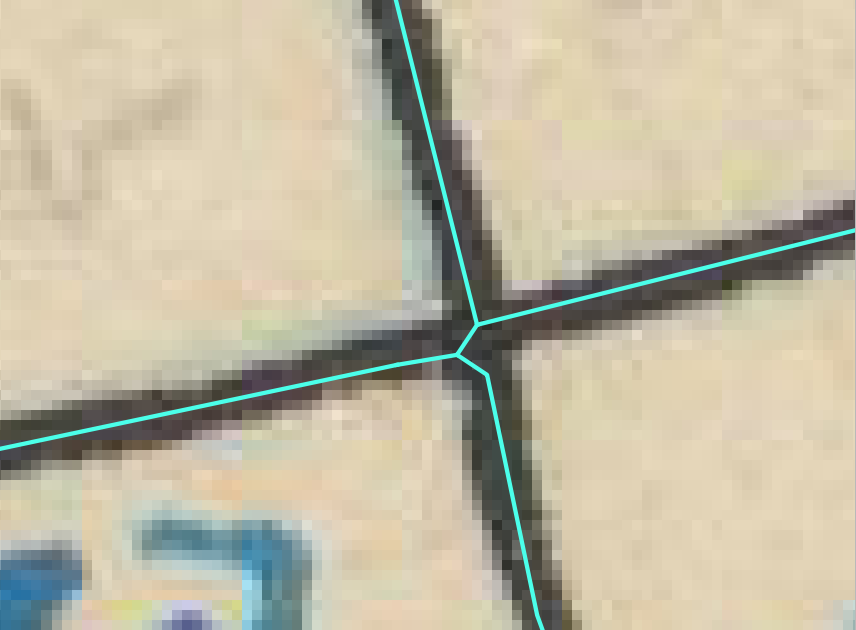}}
        \fbox{\includegraphics[width = 0.45\linewidth, height = 0.5\linewidth]{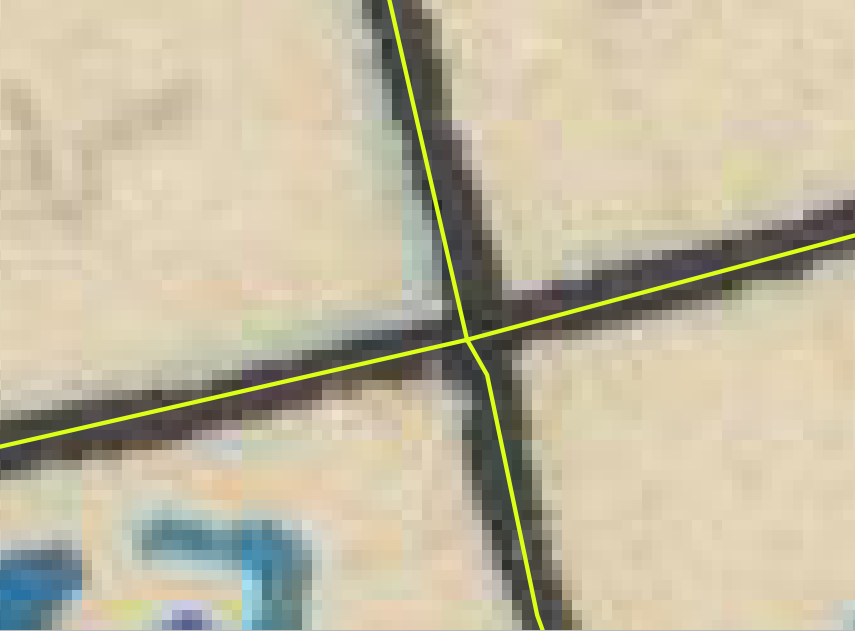}}
        \label{fig:join_error}
    }
    \hfill
    \subfloat[Un-smooth lines(left) and smooth lines(right)]{
        \fbox{\includegraphics[width = 0.45\linewidth, height = 0.5\linewidth]{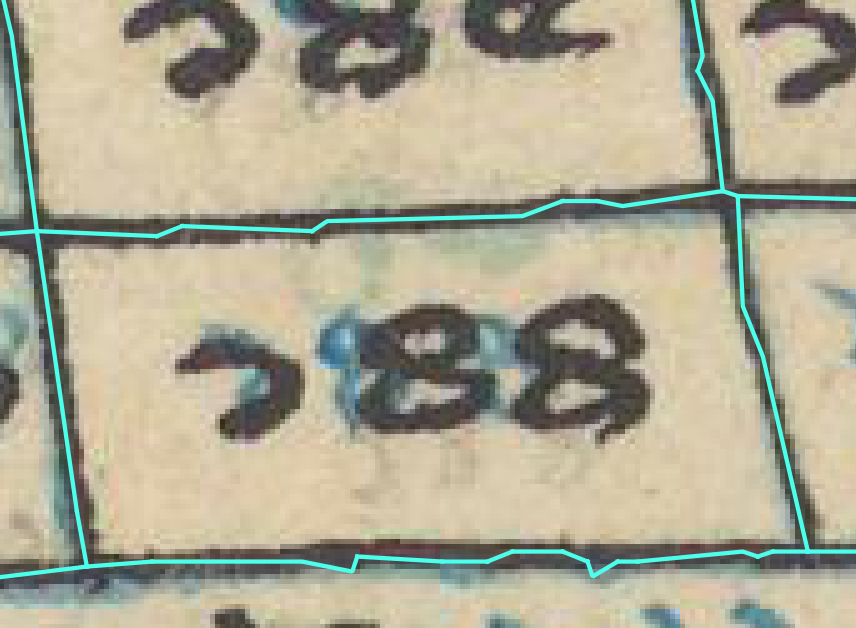}}
        \fbox{\includegraphics[width = 0.45\linewidth, height = 0.5\linewidth]{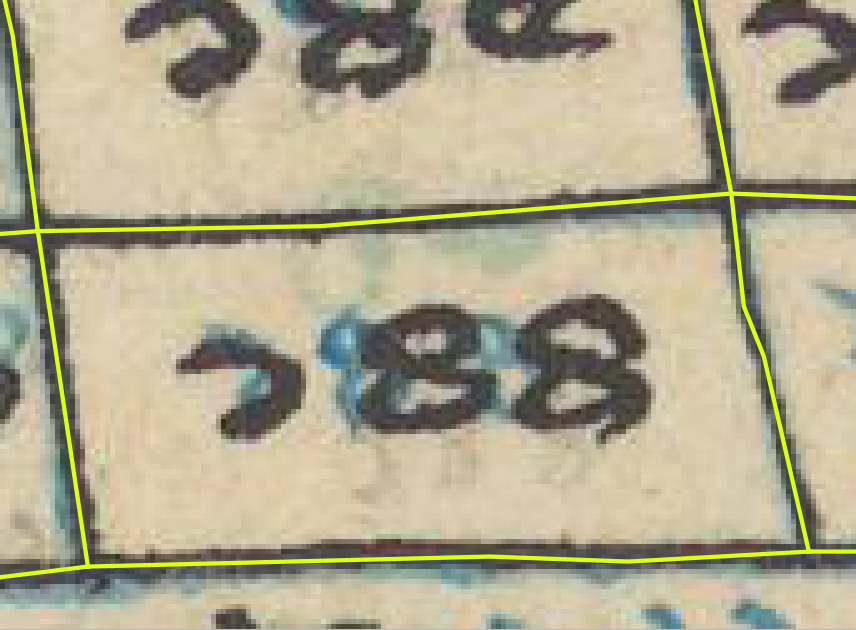}}
        \label{fig:staircase_eror}
    }
    \caption{Issues we have solved with our smoothing algorithms}
    \label{fig:grass_error}
\end{figure}

\SetKwComment{Comment}{/* }{ */}
\begin{algorithm}
\caption{Removing staircase effect}\label{alg:remove_staircase}
\KwData{C (a category)}
\KwResult{C (category) with no stair case}
\If{$len(C.points) < 4$}{
    \Return
}
$P \gets C.getPoints()$\;
$NP \gets []$ \Comment*[r]{new points, empty list}
$NP.append(P[0])$\;
$i \gets 0$
\For{$i < len(P)-3$}{
    $dir1 \gets direction(P[i],P[i+1],P[i+2])$\;
    $dir2 \gets direction(P[i+1],P[i+2],P[i+3])$\;
    \Comment{direction can be +1 or -1. for acticlock-wise +1 and clock-wise -1}
    \eIf{$dir1 \times dir2 < 0$}{
        \Comment{stair case found}
        $mp \gets getMidPoint(P[i+1],P[i+2])$
        $NP.remove(P[i+1], P[i+2])$
        $NP.appned(midPoint(mp))$\;
        $i \gets i+1$\;
    }{
         $NP.append(P[i+1])$\;
    }
}
$C.setPoints(NP)$
\end{algorithm}
\SetKwComment{Comment}{/* }{ */}
\begin{algorithm}
\caption{Removing zero length lines}\label{alg:remove_size0_line}
\KwData{categories}
\KwResult{categories with no zero length lines}
\For{$cat \in categories$}{
    $P \gets cat.getPoints()$\;
    $newPoints \gets []$  \Comment*[r]{empty list}
    $newPoints.append(P[0])$\;
    $i \gets 0$\;
    \For{$i \leq len(P)-1$}{
        \If{$getdistance(P[i],P[i+1]) \neq 0$}{
            $newPoints.append(P[i+1])$\;
        }
    }
    $cat.setPoints(newPoints)$
}
\end{algorithm}
\SetKwComment{Comment}{/* }{ */}
\begin{algorithm}
\caption{Merging categories}\label{alg:merge_categories}
\KwData{C (categories)}
\KwResult{$C_n$ (categories after merging)}
\Comment{Creating a mapping of each category by two endpoints}
$D \gets \{\}$ \Comment*[r]{empty dictionary}
\For{$cat \in C$}{
    $D[cat.leftPoint].append(cat)$\;
    $D[cat.rightPoint].append(cat)$\;
}
\For{$point \in D.keys()$}{
    \If{$len(D[point]) = 2$}{
        $cat1, cat2 \gets D[point][0,1]$\;
        $cat1.merge(cat2)$\;
        \Comment{cat2 merged into cat1}
        $D[cat2.leftPoint].replace(cat2, cat1)$\;
        \Comment{repeat for right point}
        $cat2.removed \gets true$\;
    }
}
$C_n \gets []$\Comment*[r]{empty list}
\For{$cat \in C$}{
    \If{$not cat.removed$}{
    $C_n.append(cat)$\;
    }
}
\Return $C_n$\;
\end{algorithm}
Grass GIS \cite{GRASS_GIS_software} output has many flaws that must be addressed. We tried to solve the most frequent problems such as multiple line joining at a point problem and unsmooth lines which occur because of the thinning process of Grass GIS. \cref{fig:grass_error} illustrates these problems.

\textbf{Solving Join Error.} We have implemented \cref{alg:correct_join} to address the join error. The issue and its resolution are demonstrated in \cref{fig:join_error}. This algorithm examines all such categories illustrated in \cref{fig:ascii}. The algorithm first identifies instances of this error by searching for lines connected to four different categories—two on each side. When such a configuration is detected, we check whether the length of the line is sufficiently small than a threshold value. If the line's length is below this threshold, indicating it is sufficiently small, the corresponding category is marked. This marked category consists of two points and represents that particular line. Subsequently, the algorithm connects these four categories to the midpoint of the marked category and removes the marked category.

\textbf{Remove Staircase Effect}. We observe most of the vector lines have a zigzag pattern, which we refer to as the staircase effect (in \cref{fig:staircase_eror}). To remove this, we implement \cref{alg:remove_staircase}. To identify this effect, we consider four consecutive points from a category and examine whether the angle of rotation of the first three is opposite to the angle of rotation of the last three, illustrated in \cref{fig:staircase}. Upon detecting such a pattern, the algorithm replaces the middle two points with their midpoint. This process is iteratively applied to consecutive sets of four points, effectively smoothing out the zigzag pattern.

\usetikzlibrary{calc, patterns, angles, quotes}
\begin{figure}
    \subfloat[]{
        \fbox{
        \begin{tikzpicture}[scale=0.9]
      \coordinate [label=above:\scriptsize$p1$] (p1) at (1,1);
      \coordinate [label=below:\scriptsize$p2$] (p2) at (1.8,1);
      \coordinate [label=above:\scriptsize$p3$] (p3) at (2.1,1.5);
      \coordinate [label=above:\scriptsize$p4$] (p4) at (2.9,1.5);

      \filldraw[black](p1) circle (1pt);
      \filldraw[black](p2) circle (1pt);
      \filldraw[black](p3) circle (1pt);
      \filldraw[black](p4) circle (1pt);
      \draw (p1)--(p2);
      \draw (p2)--(p3);
      \draw (p3)--(p4);

      \end{tikzpicture}
        }
    }
    \subfloat[]{
        \fbox{
        \begin{tikzpicture}[scale=0.9]
          \coordinate [label=above:\scriptsize$p1$] (p1) at (1,1);
          \coordinate [label=below:\scriptsize$p2$] (p2) at (1.8,1);
          \coordinate [label=above:\scriptsize$p3$] (p3) at (2.1,1.5);
          \coordinate [label=above:\scriptsize$p4$] (p4) at (2.9,1.5);
    
          \filldraw[black](p1) circle (1pt);
          \filldraw[black](p2) circle (1pt);
          \filldraw[black](p3) circle (1pt);
          \filldraw[black](p4) circle (1pt);
          \draw (p1)--(p2);
          \draw (p2)--(p3);
          \draw (p3)--(p4);
          
          \pic [draw=red, text=blue, ->, "$+ve$", angle eccentricity=1.5] {angle =p3--p2--p1};

          \pic [draw=red, text=blue, <-, "$-ve$", angle eccentricity=1.5] {angle =p2--p3--p4};
          \end{tikzpicture}
        }
    }
    \subfloat[]{
        \fbox{
        \begin{tikzpicture}[scale=0.9]
          \coordinate [label=above:\scriptsize$p1$] (p1) at (1,1);
          \coordinate [label=below:\scriptsize$p2$] (p2) at (1.8,1);
          \coordinate [label=above:\scriptsize$p3$] (p3) at (2.1,1.5);
          \coordinate [label=above:\scriptsize$p4$] (p4) at (2.9,1.5);
          \coordinate [label=right:\scriptsize] (mid) at (1.95,1.25);
    
          \filldraw[black](p1) circle (1pt);
          \filldraw[black](p2) circle (1pt);
          \filldraw[black](p3) circle (1pt);
          \filldraw[black](p4) circle (1pt);
          \filldraw[blue](mid) circle (1pt);
          \draw (p1)--(p2);
          \draw (p2)--(p3);
          \draw (p3)--(p4);

          \draw[blue] (p1)--(mid)--(p4);
          \end{tikzpicture}
        }
    }
    \caption{a) Staircase effect with four points b) Detecting this issue by checking the opposite angle of rotation c) Solving the issue by passing the line through mid-point}
    \vspace{-10pt}
    \label{fig:staircase}
\end{figure}
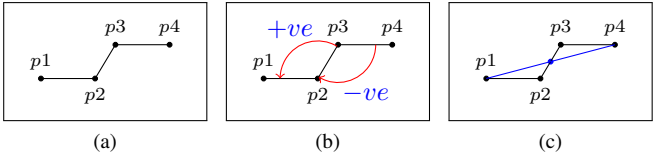
In addition to the aforementioned \cref{alg:correct_join} and \cref{alg:remove_staircase} the removal of zero-length categories (\cref{alg:remove_size0_line}) and the merging of certain categories (\cref{alg:merge_categories}) are essential to simplify the smoothing process.
\subsection{Integrate Plots with corresponding Plot Numbers}
\label{subsec:join_plot_numbers}
Here we polygonize the vector file with QGIS's \cite{QGIS_software} functionality. Each polygon has a polygon ID. Now, from the digit-position pair we had from \cref{subsec:digit_recognition}, for each digit, we check whether its position falls within a specific polygon. By iterating through all the digits, we assign each digit to the polygon that contains its coordinates. This results in grouping all digits according to their respective polygons.

We have observed that the plot numbers, in the mouza map, exhibit a consistent spatial orientation, which is, that the most significant bit (MSB) consistently resides on the left side, while the least significant bit (LSB) consistently occupies the right side of the map. Simply, the plot numbers are written left to right. This orientation helps to generate the plot number within the polygon. After getting each digit in their respective polygon we scan the digits from left to right in the X-axis or simply sort them according to their increasing X-axis value and generate the plot number. For example, consider the scenario where we have identified the digits $2$ at coordinates $(2,2)$, $9$ at coordinates $(1,2)$, and $5$ at coordinates $(-1,2)$ within a given polygon. Employing a left-to-right sorting approach along the X-axis, we generate the plot number $592$. Then we join plot numbers associated with their respective polygons in the vector database, completing the full vectorization of the en map.

This phase of the process is executed through the QGIS \cite{QGIS_software} python plugin, contributing to the automation of our pipeline. While we utilized the QGIS Python plugin for polygonization due to its robust geometric and coordinate handling functionalities, the process of generating and integrating digitized plot numbers into the dataset was developed independently by us. Significantly, the entire workflow is accomplished without any human intervention. The combination of these automated steps (from \cref{susubsec:boundary_inpainting}) ensures the generation of a vectorized mouza map in a single, continuous operation.
\section{Experimental Evaluation}
\label{sec:experimental_evaluation}
In this section, we evaluate our methodology both qualitatively and quantitatively on different maps by presenting a comparative analysis against manual vectorization.

\begin{figure*}
    \subfloat[]{
        \fbox{\includegraphics[width = \textwidth, height = 0.48\textheight]{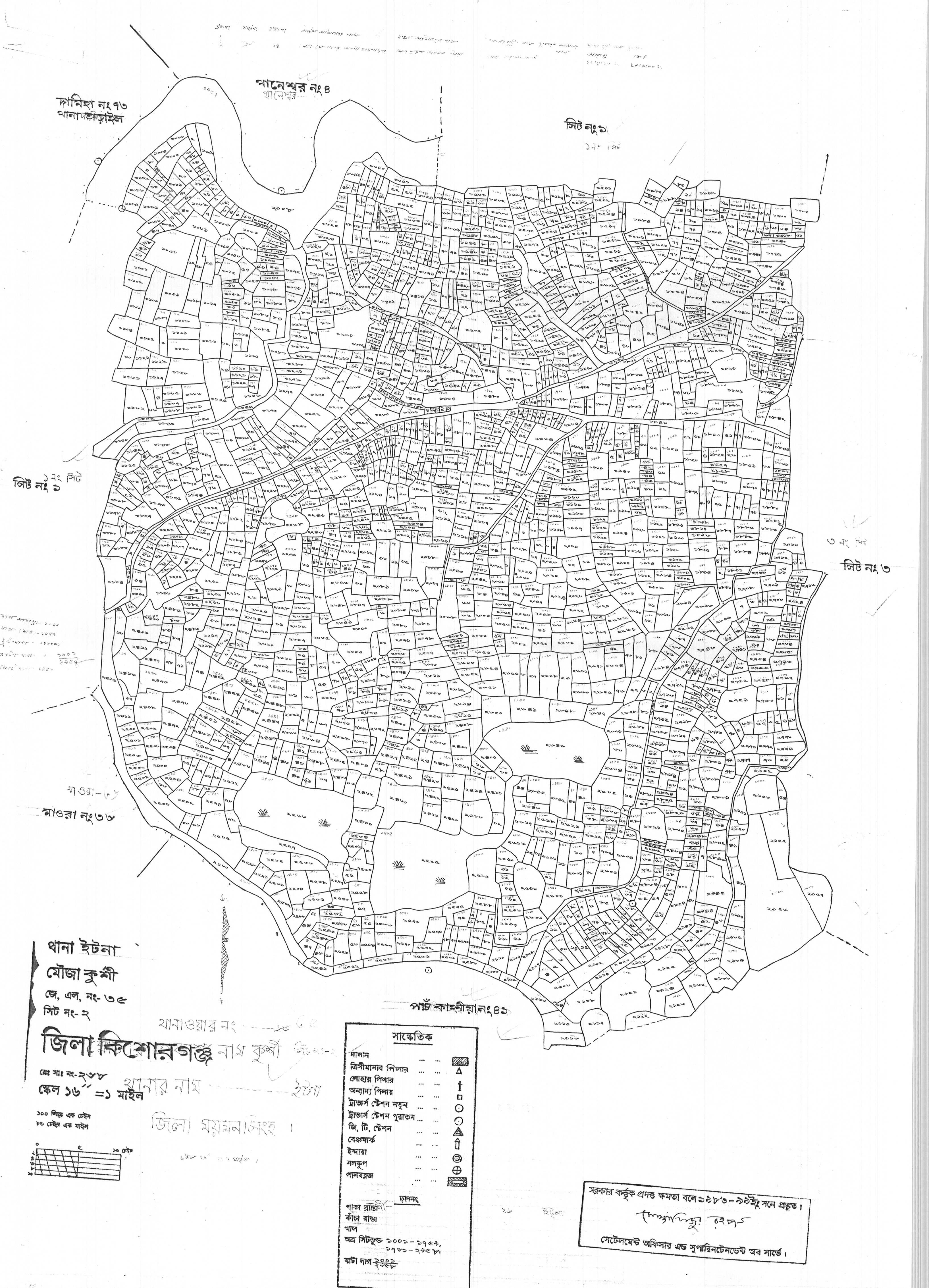}}
    } \\
    \\
    \subfloat[]{
        \fbox{\includegraphics[width = 0.48\textwidth, height = 0.4\textheight]{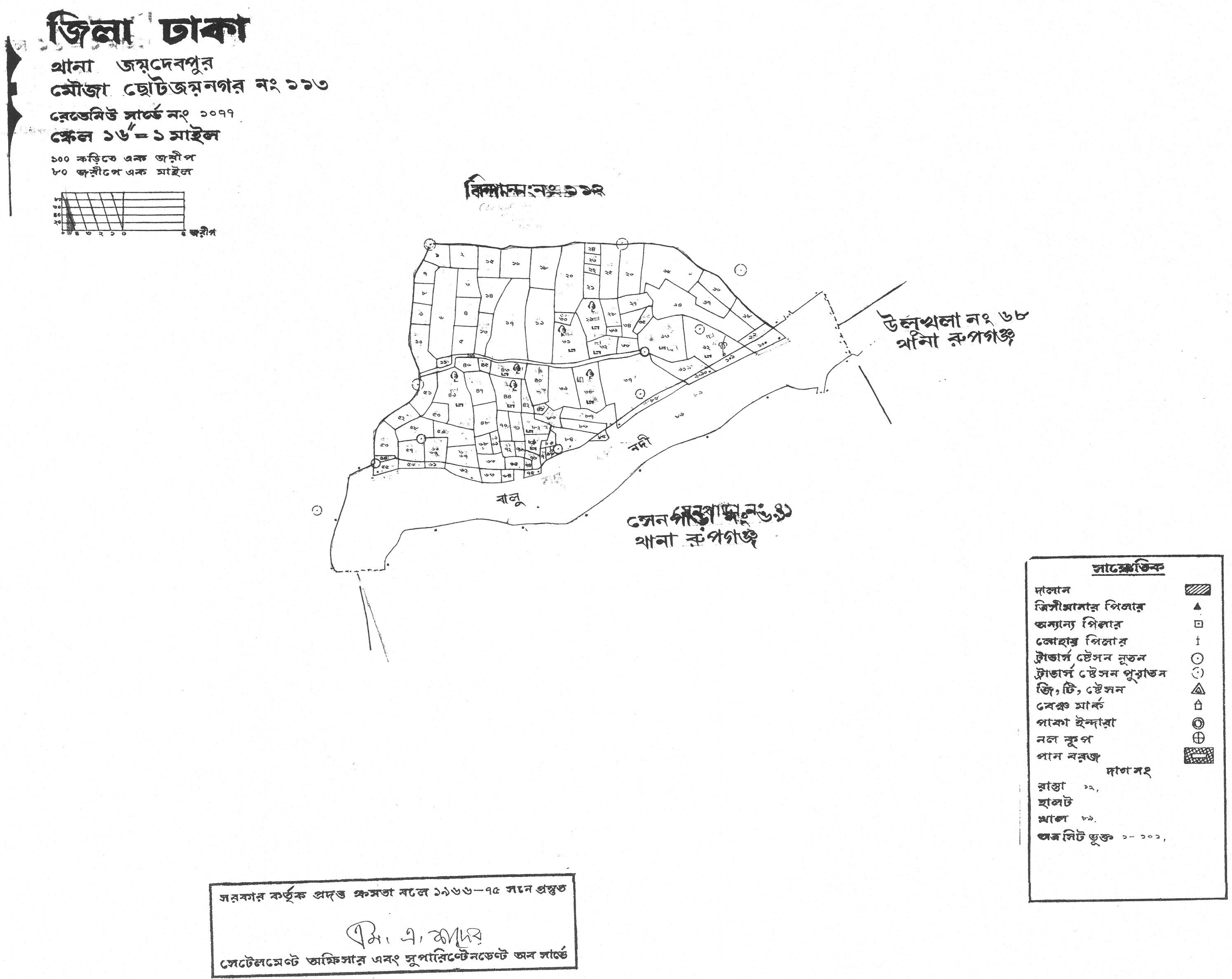}}
    } \hspace{2pt}
    \subfloat[]{
        \fbox{\includegraphics[width = 0.485\textwidth, height = 0.4\textheight]{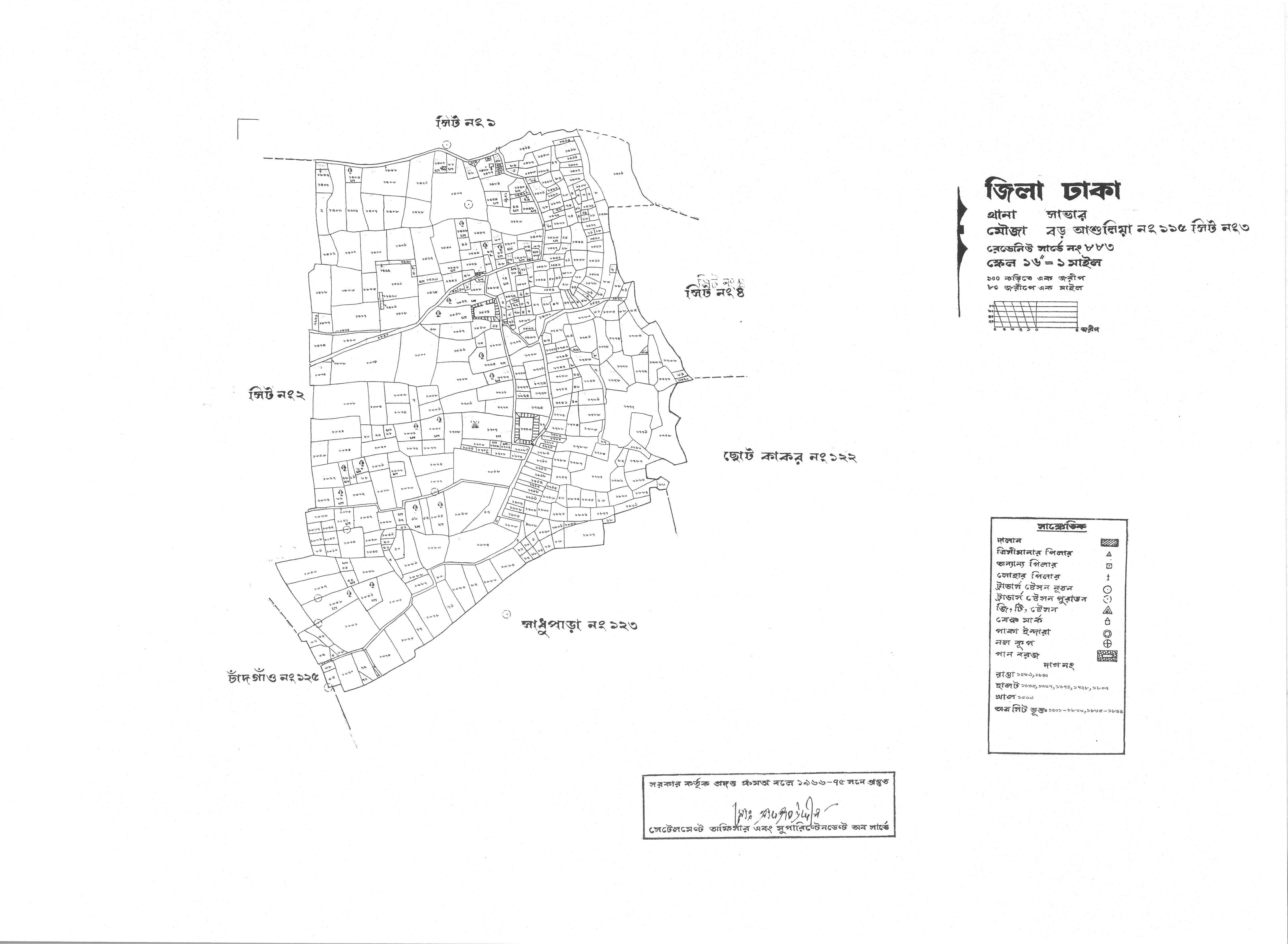}}\hspace*{-2pt}
    }
    
    \caption{Three types of maps we have performed our experiment on. a) Difficult type, number of plots: 1957 b) Easy type, number of plots: 101 c) Medium type, number of plots: 441}
    \vspace{-10pt}
    \label{fig:map_types}
\end{figure*}

\subsection{Experimental Setup}
\label{subsec:exp_setup}
Our experiments were conducted on two Windows machines, one with 16 GB RAM and an AMD Ryzen 5 3600 6-Core processor @3.59GHz, and the other with 8 GB RAM and an Intel(R) Core(TM) i7 CPU @1.80GHz. Coding was executed in \emph{Python}, utilizing \emph{Google Colab} for training the line joining and digit predictor models with \emph{Pytorch}. Image processing relied on \emph{OpenCV}, while raster image editing utilized \emph{Microsoft Paint} and \emph{Paint 3D}. For vectorization, we used \emph{GRASS GIS 8.0} and for joining plot numbers \emph{QGIS 3.24.3} was utilized.

\textbf{Datasets.} We collect a set of scanned image files from the Land Office of Bangladesh\cite{landzone}. These images are also accessible to institutions involved in the manual vectorization of Mouza maps. In addition to the scanned images, we have collected our experimental data from these institutions. Subsequently, we conducted our experiments by running our code on personal computers. During these experiments, we meticulously recorded the time required for each process. This comprehensive data collection process formed the foundation of our research.

\begin{figure*}
    \fbox{\includegraphics[width = 0.24\textwidth, height = 0.25\linewidth]{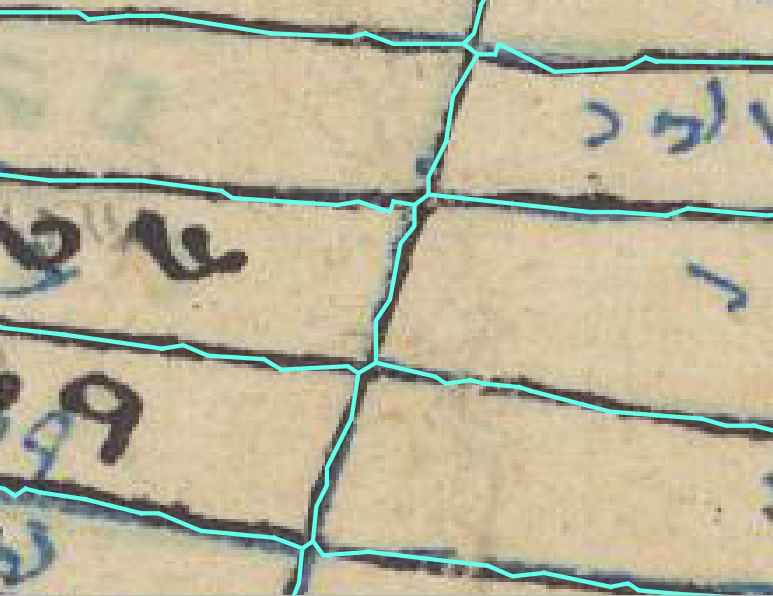}}\hspace*{-2pt}
    \fbox{\includegraphics[width = 0.24\textwidth, height = 0.25\linewidth]{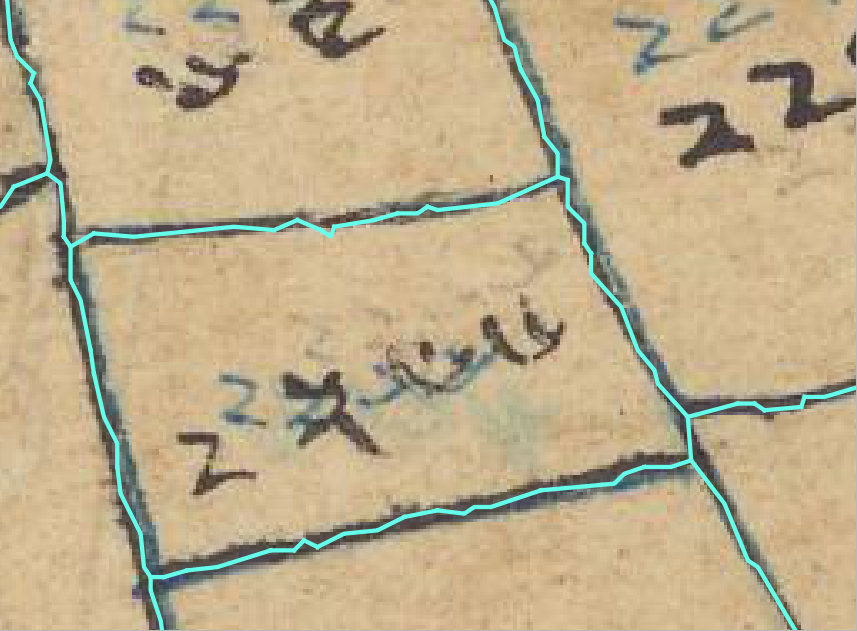}}\hspace*{-2pt}
    \fbox{\includegraphics[width = 0.24\textwidth, height = 0.25\linewidth]{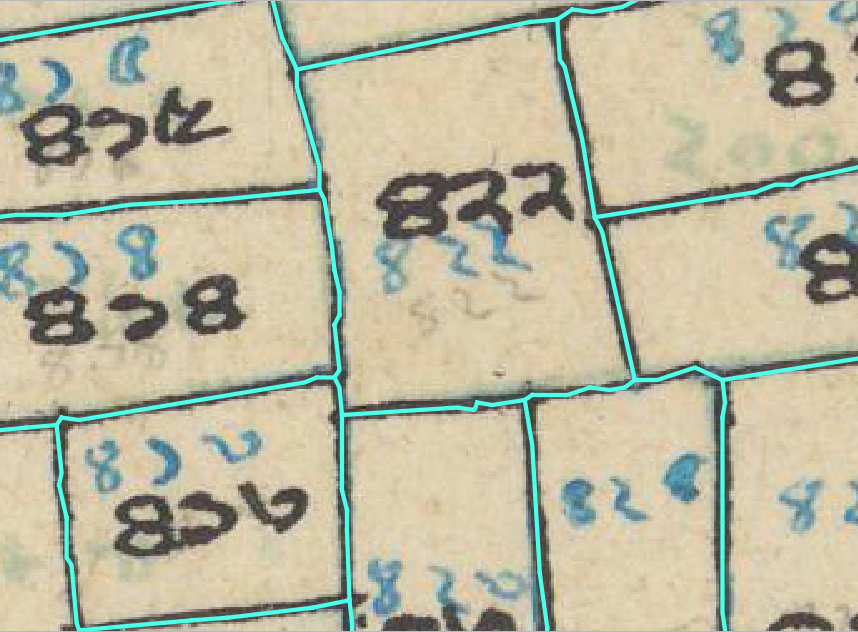}}\hspace*{-2pt}
    \fbox{\includegraphics[width = 0.24\textwidth, height = 0.25\linewidth]{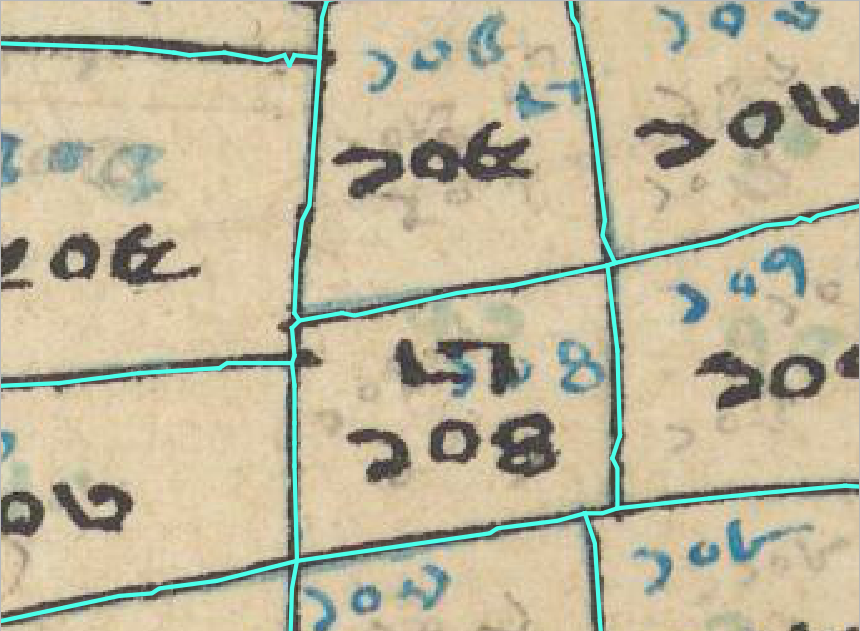}}
    \\
    \fbox{\includegraphics[width = 0.24\textwidth, height = 0.25\linewidth]{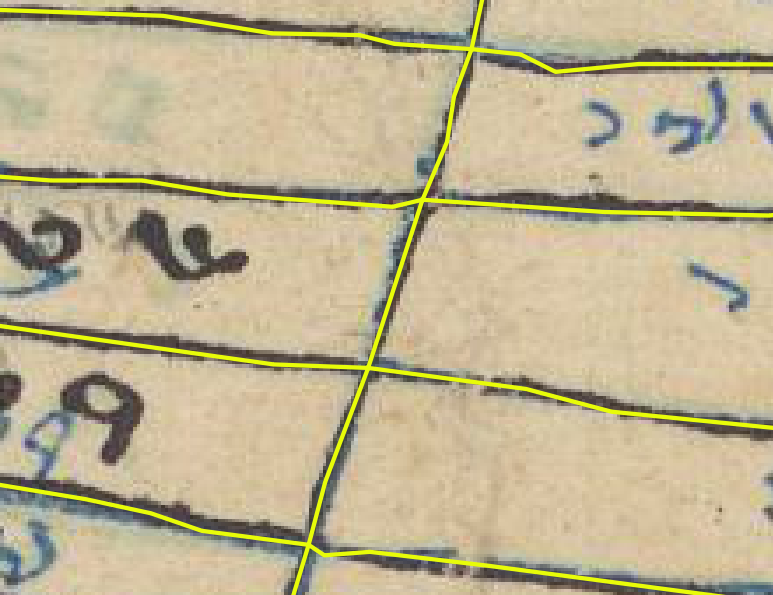}}\hspace*{-2pt}
    \fbox{\includegraphics[width = 0.24\textwidth, height = 0.25\linewidth]{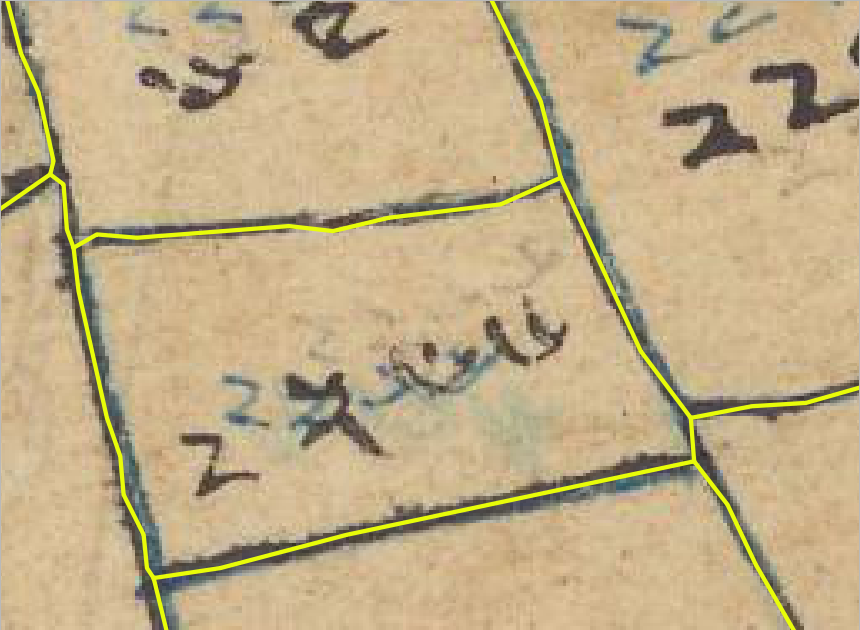}}\hspace*{-2pt}
    \fbox{\includegraphics[width = 0.24\textwidth, height = 0.25\linewidth]{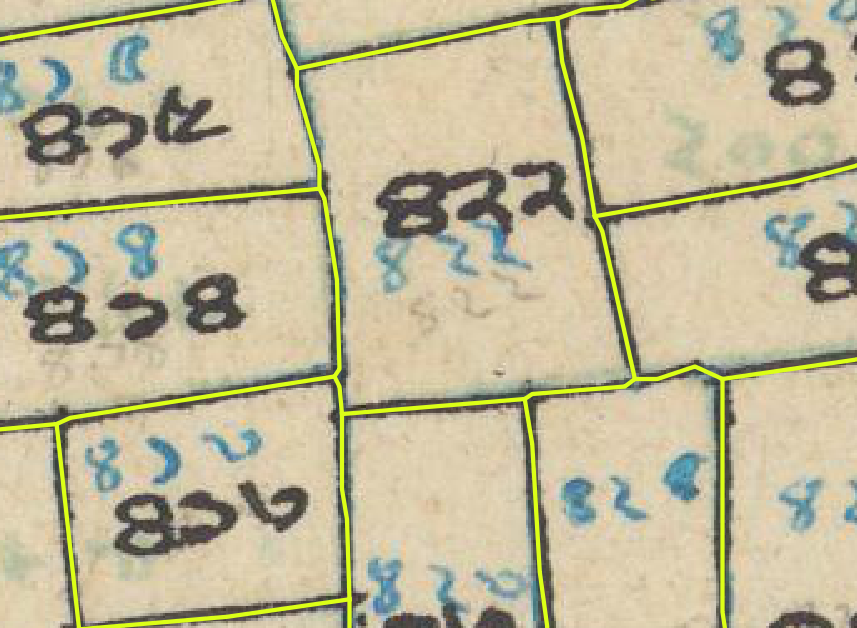}}\hspace*{-2pt}
    \fbox{\includegraphics[width = 0.24\textwidth, height = 0.25\linewidth]{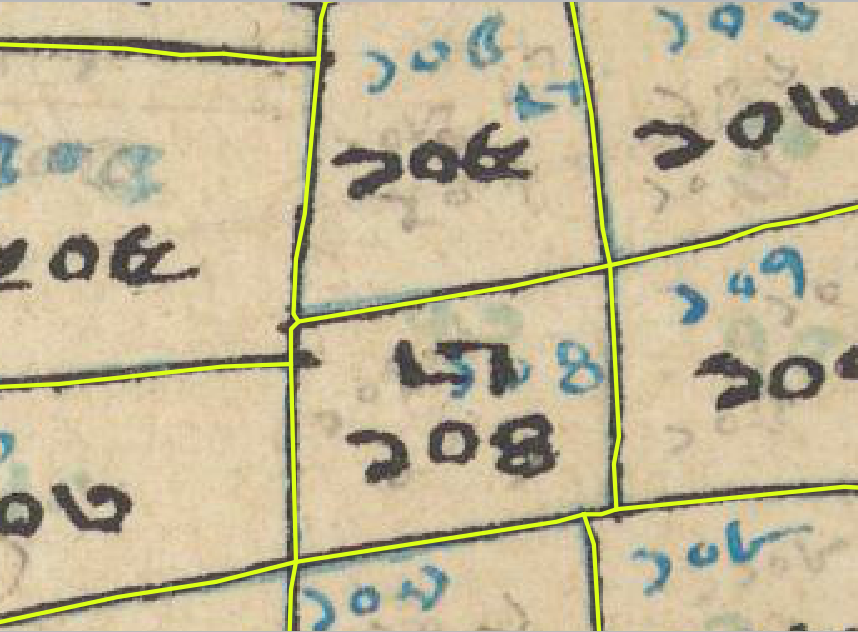}}
    \\
    \fbox{\includegraphics[width = 0.24\textwidth, height = 0.25\linewidth]{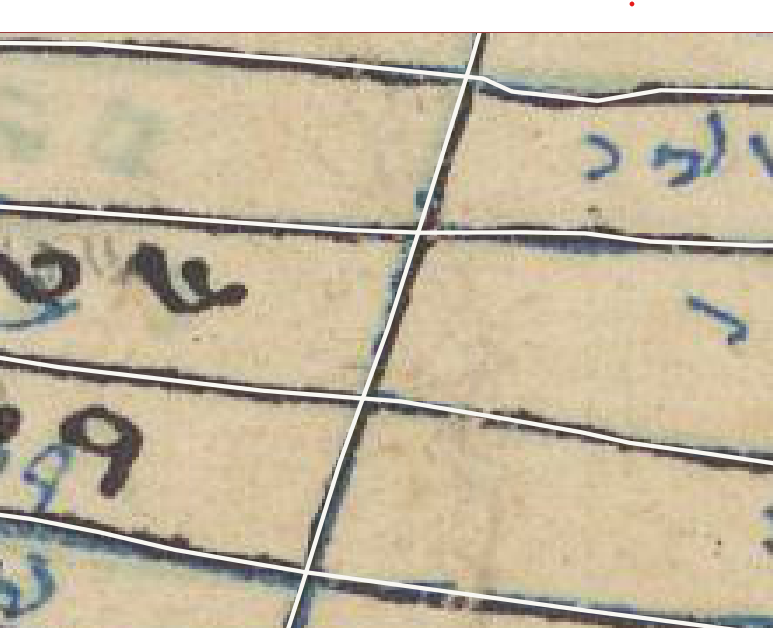}}\hspace*{-2pt}
    \fbox{\includegraphics[width = 0.24\textwidth, height = 0.25\linewidth]{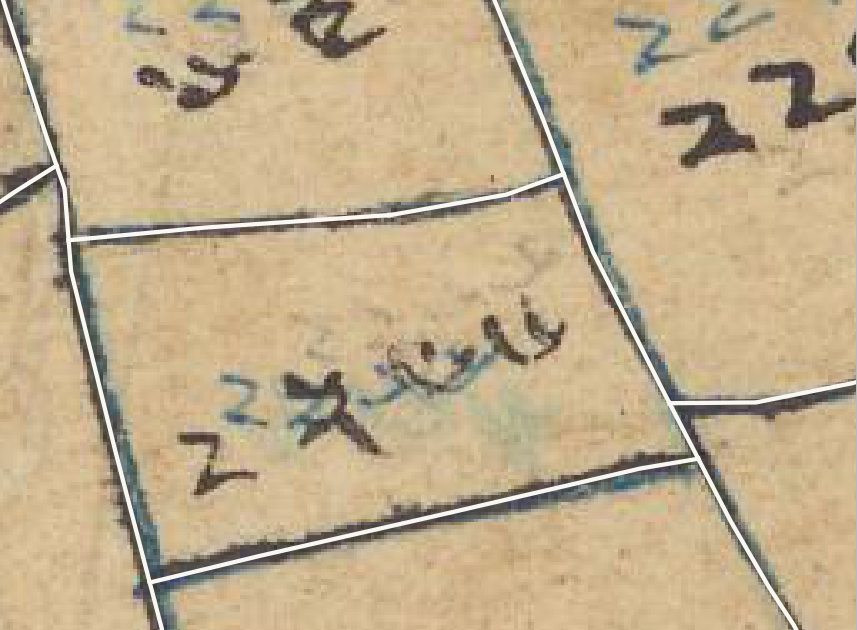}}\hspace*{-2pt}
    \fbox{\includegraphics[width = 0.24\textwidth, height = 0.25\linewidth]{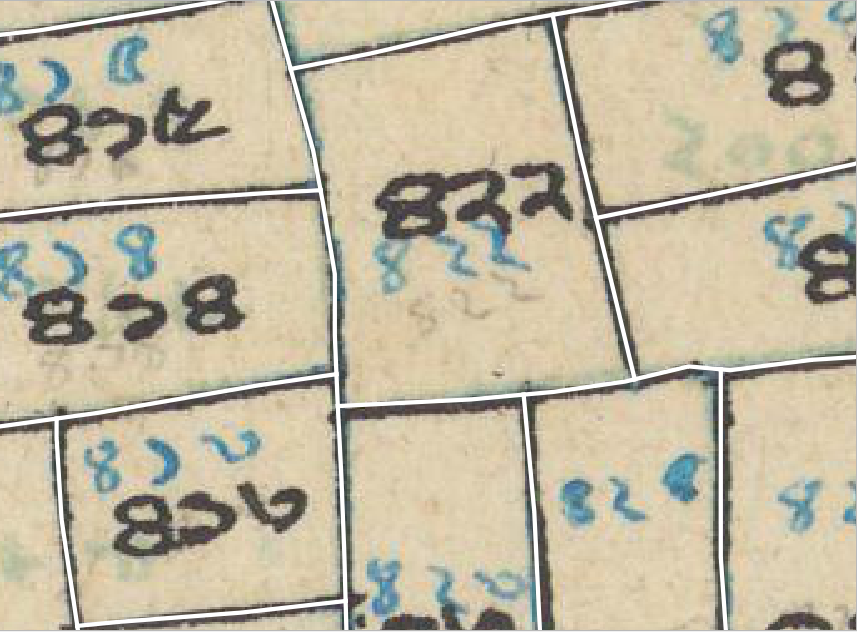}}\hspace*{-2pt}
    \fbox{\includegraphics[width = 0.24\textwidth, height = 0.25\linewidth]{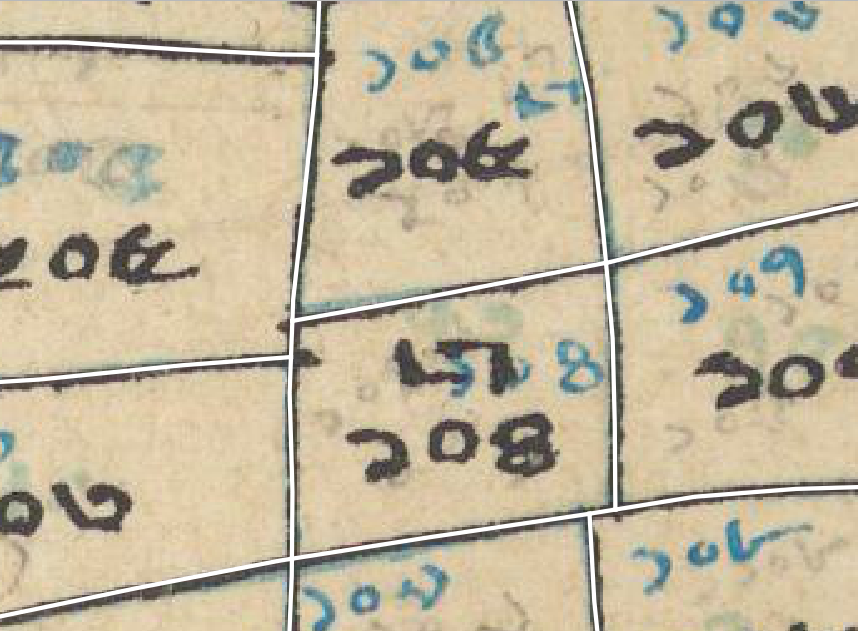}}
    
    \caption{1) \textbf{first row}: blue lines are \textbf{GRASS GIS} outputs. 2) \textbf{second row}: yellow lines are \textbf{ours} method output. 3) \textbf{third row}: white lines are outputs of \textbf{manual} vectorization (ground truth)}
    \vspace{-10pt}
    \label{fig:all_outputs}
\end{figure*}

\subsection{Experiment Results}
\label{subsec:exp_result}
We have completed our whole experiments on eight maps which in total contain approximately 4800 plots. This extensive number of plots allows a comprehensive evaluation of our proposed methodology. Some portions of different maps are shown in \cref{fig:all_outputs} for a qualitative comparison, where colored vector lines are overlaid on their raster counterparts. We can see, that our output does not perfectly match the ground truth but it closely resembles it.

Our experimental process is divided into three sections. The first stage of the process involves the manual editing of the raster maps which involves pre-processing, boundary inpainting, and connected component analysis. Then, in the second part of the process, we run the code which covers the rest of the methodology including plot number detection, vectorization, smoothing, and merging plot numbers in the vector database. The output of our methodology is not precise as we can see in \cref{fig:all_outputs}, a requisite level of post-processing refinement is essential to attain the desired quality in the final output. We refer to this refinement part as vector editing. The time log for each component is outlined in \cref{tab:time_log}. Map difficulty is evaluated based on the plot numbers, plot areas, and the intricate structure of map boundaries (see \cref{fig:map_types}).

\begin{table}[h]
 \caption{Time Log (In minutes)}
    \label{tab:time_log}
    \centering
    \setlength\tabcolsep{3pt}
    \begin{tabular}{|c| c| c| c| c |}
        \hline
       \multicolumn{1}{|c| }{Map} &
       \multirow{2}{*}{Plots} & 
       \multicolumn{1}{c| }{Raster } & 
       \multicolumn{1}{c| }{Code  } & 
       \multicolumn{1}{c|}{Vector }\\
       \multicolumn{1}{|c| }{type} &&
       \multicolumn{1}{c| }{editing} & 
       \multicolumn{1}{c| }{run} & 
       \multicolumn{1}{c|}{editing}\\
        \hline
        {Easy} & {101} & {5} & {2} & {82}\\ \hline
        {Easy} & {254} & {8} & {2} & {187}\\ \hline
        {Easy} & {301} & {10} &{2} & {250}\\ \hline
        {Medium} & {441} & {12} & {2} & {310}\\ \hline
        {Difficult} & {423} & {20} & {2} & {390}\\ \hline
        {Difficult} & {496} & {26} & {3} & {425}\\ \hline
        {Difficult} & {849} & {43} & {3} & {550}\\ \hline
        {Difficult} & {1957} & {60} & {4} & {720}\\
        \hline
    \end{tabular}
   
\end{table}

    

It is evident that a medium-sized map containing 441 plots requires less editing time compared to a challenging map with 423 plots. This observation highlights that the refinement processes, both for raster and vector components, are influenced not only by the number of plots but also by factors such as map dimensions and plot contours.

Finally, we examine the time allocation for different stages of the process. Notably, the code execution phase accounts for less than 1\% of the total time, whereas the most substantial portion of time, approximately 94\%, is dedicated to vector editing. The remaining raster editing incurs a significant time investment; however, this time duration remains distinctly distant from vector editing.

\subsection{Time Comparison}
\label{subsec:comparison}
\begin{table}[h]
     \caption{Time (In hours) comparison between repairing our output with manual vectorizing for different types of maps.}
    \label{tab:comparison}
    \centering
    \begin{tabular}{|c|c|c|c|c|c|c|}
        \hline
       \multirow{2}{*}{Map type} & 
       \multirow{2}{*}{Plot} &
       \multirow{2}{*}{Our (Hours)} & 
       \multicolumn{2}{c|}{Manually (Hours) }&
       \multicolumn{2}{c|}{Saving(\%)}\\
       \cline{4-7}
       &&&Average&Best&Min&Max\\
        \hline
        {Easy}&{101}&{1.48}& {3} &{2} &{26} &{50.66}\\
        \hline
        {Easy}&{254}&{3.38}& {5} &{4} &{15.5} &{32.4}\\
        \hline
        {Easy}&{301}&{3.86}& {5.5} &{4.5} &{14.2} &{29.8}\\
        \hline
        {Medium}&{441}&{5.4}& {9} &{7.5} &{28} &{40}\\
        \hline
        {Difficult}&{423}&{6.86}& {11} &{9.5} &{27.78} &{37.64}\\
        \hline
        {Difficult}&{496}&{7.56}& {14} &{10} &{24.4} &{46}\\
        \hline
        {Difficult}&{849}&{9.93}& {20} &{16} &{37.94} &{50.35}\\
        \hline
        {Difficult}&{1957}&{13}& {25} &{20} &{35} &{48}\\
        \hline
    \end{tabular}
\end{table}
In this section, we present time comparisons between the established procedures for generating a vector map from inception and the process of refining our generated vector map, as outlined in \cref{tab:time_log}. Time estimates for the manual procedure were sourced from domain experts. To validate the efficacy of our editing approach, we engaged in collaboration with an organization specializing in manual vectorization. Qualified experts conducted a thorough verification of our editing process and furnished time estimates pertaining to manual vectorization.

The results of the manual processes are detailed in two columns of \cref{tab:comparison}, one representing average case time and the other, the best case scenario involving a proficient operator engaged in manual vectorization of mouza maps. In our observations across all maps, it is evident that the time required for editing our maps is notably less compared to the best-case scenario of manual editing.

\begin{figure}[h]
    \centering
    \begin{tikzpicture}
\begin{axis}[
scaled y ticks=real:1,
ytick scale label code/.code={},
ymax = 30,
symbolic x coords={101,254,301,423,441,496,849,1957},
xtick=data,
height=7cm,
width=8cm,
grid=major,
xlabel={Number of plots},
ylabel={Time (Hours)},
legend style={
cells={anchor=east},
legend pos=north west,
}
]
\addplot coordinates {(101, 1.48) (254,3.38) (301, 3.86) (423, 6.86) (441, 5.4) (496, 7.56) (849, 9.93) (1957, 13)};
\addplot coordinates {(101, 3) (254,5) (301, 5.5) (423, 11) (441, 9) (496, 14) (849, 20) (1957, 25)};
\addplot coordinates {(101, 2) (254,4) (301, 4.5) (423, 9.5) (441, 7.5) (496, 10) (849, 16) (1957, 20)};

\legend{Our (Hours), Manually average (Hours), Manually best (Hours)}
\end{axis}

\end{tikzpicture}
    \caption{Time vs Number of plots}
    \vspace{-10pt}
    \label{fig:plot_chart}
\end{figure}
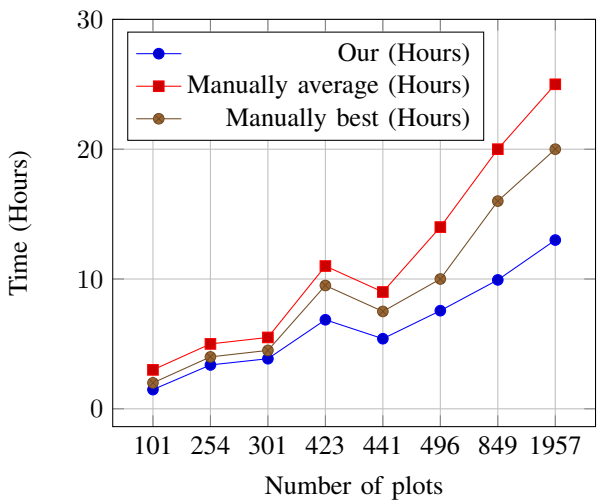

We calculate the percentage of the time, that is saved, which is for an average case of $32-50\%$ of time. Even if an efficient operator does the manual vectorization, our process will save $14-37\%$ of time. We have completed our process on approximately 4800 mouza plots. In our comprehensive comparative analysis, we bring attention to a substantial discovery. As depicted in \cref{fig:plot_chart}, the slope of our methodology stands out for its diminutive magnitude in comparison to alternative approaches. This observation suggests that our methodology excels, particularly in handling larger or more challenging maps.

\subsection{Vector Comparison}
\label{subsec:quantitative_comparison}
Our research focuses on the complete digitization of Mouza maps, addressing several complex aspects. In this section, we specifically compare the vectorization output of plot boundaries, which is described in \cref{subsec:vectorization}, against the results acquired from the Deep Vectorization\cite{rassian}. We assess the performance of these vectorization techniques using various metrics—Intersection over Union (IoU), Hausdorff Distance, Frechet Distance, and Mean Squared Error (MSE)—which effectively capture the artifacts and accuracy of each method.\\

\textbf{Intersection-over-Union (IoU)}\cite{IOU} measures how much two raster shapes or rasterized vector drawings overlap. However, IoU does not effectively measure differences between shapes that are similar but slightly shifted in position. IoU between two shapes \( R_1 \) and \( R_2 \) is given by:\\
\[ \text{IoU}(R_1, R_2) = \frac{|R_1 \cap R_2|}{|R_1 \cup R_2|}\]

\textbf{Hausdorff Distance}\cite{hausdorff_wikipedia} measures the deviation of two shapes even if they are similar. It quantifies the greatest deviation of two shapes.The Hausdorff distance \( d_H \) between two sets of points \( M \) and \( N \) is defined as:

\[
d_H(M, N) = \max \left\{ \sup_{m \in M} \inf_{n \in N} d(m, n), \sup_{n \in N} \inf_{m \in M} d(n, m) \right\}
\]

\textbf{The Discrete Frechet Distance}\cite{frechet} measures the similarity between two curves. If there are two curves P and Q parameterized by p(t) and q(t) respectively where t is a parameter ranging from 0 to 1:
\[
d_F(P,Q) = \inf_{p, q} \max_{t \in [0,1]} \| P(p(t)) - Q(q(t)) \|
\]
Where (1)$\|\cdot \|$  indicates the Euclidean distance between points, and  (2) $p(t)$  and  $q(t)$  are continuous and non-decreasing reparameterizations of the curves P  and  Q.

   

\begin{table}[h]
     \caption{Comparison of IOU, Hausdorff, Frechet Distance and MSE}
    \label{tab:metrics}
    \centering
    \begin{tabular}{|c|c|c|}
        \hline
         \multirow{2}{*}{Method} &  \multirow{2}{*}{Deep}&  \multirow{2}{*}{Ours}\\
          & &  \\
        
        \hline
         {IOU,\%} & {51.04} & {\textbf{63.39}}\\
         \hline
         {Hausdorff} & {19.36} &{\textbf{15.72}}\\
         \hline
         {Frechet Distance} & {24.96} &{\textbf{16.21}}\\
         \hline
         {MSE,\%} &{4.11}&{\textbf{2.09}}\\
        
        \hline
    \end{tabular}
   
\end{table}

For evaluation, we have collected the ground truth data from the institutions mentioned in the dataset section, which are involved in the manual digitization of mouza maps. Given that, the raster maps are typically in the range of $8000\times10000$ pixels, we have created $512\times512$ patches from several images and evaluated them on approximately $1000$ patches except for the frechet distance. For frechet distance, we used $64\times64$ patches and evaluated approximately $3000$ of them. We present the evaluation results in \cref{tab:metrics}. It can be observed that both the Hausdorff distance and the Discrete Frechet distance are lower in our method, indicating the deviation of the lines between the ground truth and our vectorization is minimal. Our vectorization process outperforms all the metrics. Besides the improvements, our process offers a significantly faster output. Deep vectorizer takes hours to vectorize a map using NVIDIA GeForce GTX 1660 Super GPU whereas our method takes 2-4 minutes (mentioned in \cref{tab:time_log}) to vectorize a map. Our approach is more time-efficient and less resource-intensive, providing a clear advantage in terms of both processing time and computational efficiency.

\section{Discussion}
\label{sec:discussion}
This study introduces a novel semi-automated approach for Mouza map digitization, addressing key limitations in previous works. This study is the first to incorporate an inpaint model along with binarization and noise removal in the pre-processing step. This step is very effective in reconstructing the missing regions of the maps. Moreover, the integration of OCR to detect plot numbers and merge them with plot boundaries after digitization is a significant advancement toward the digitization of Mouza maps. The goal of our methodology is to quickly complete the digitization process maintaining accuracy.

\subsection{Comparison With Related Studies}
\label{subsec:comparison_with_related_studies}
\begin{table*}[h]
    \caption{Qualitative Comparison of Key Features Across Related Research Studies}
    \label{tab:metrics}
    \centering
    \begin{tabular}{|c|c|c|c|c|c|c|c|c|}
        \hline
         \multirow{3}{*}{\textbf{Research Study}} &  \multirow{3}{*}{\textbf{Tools}} &  \multirow{3}{1.5cm}{\textbf{Pre-process}} &  \multirow{3}{*}{\textbf{OCR}} &  \multirow{3}{*}{\textbf{Vecotrization}} &  \multirow{3}{*}{\textbf{Smoothing}} &  \multirow{3}{1.5cm}{\textbf{Deployment Effort}}&  \multirow{3}{1.65cm}{\textbf{Computational Cost}} & \multirow{3}{*}{\textbf{Automation}} \\
          & & & & & & & & \\
          & & & & & & & & \\
        
        \hline
         \multirow{2}{*}{Vasile et al., \cite{vasile_map_digitization} } & \multirow{2}{*}{AutoCad} & \multirow{2}{*}{None} & \multirow{2}{*}{\xmark} & \multirow{2}{*}{Only Territory} & \multirow{2}{*}{\xmark} & \multirow{2}{*}{Low} & \multirow{2}{*}{Low} & \multirow{2}{*}{None}\\
         & & & & & & & &  \\

         \hline
         \multirow{3}{*}{Piskinaite et al., \cite{lithuanian_land} } & \multirow{3}{*}{GIS} & \multirow{3}{1.5cm}{Georeferencing, GCP} & \multirow{3}{*}{\xmark} & \multirow{3}{*}{Manual} & \multirow{3}{*}{\xmark} & \multirow{3}{*}{Low} & \multirow{3}{*}{Low} & \multirow{3}{*}{None}\\
         & & & & & & & &  \\
         & & & & & & & & \\

         \hline
         \multirow{3}{*}{Akter et al., \cite{east_delta_mouza} } & \multirow{3}{*}{Auto-Cad} & \multirow{3}{1.5cm}{  Canny edge detection} & \multirow{3}{*}{\xmark} & \multirow{3}{1.5cm}{Sweep Line Algorithm} & \multirow{3}{*}{\xmark} & \multirow{3}{*}{Low} & \multirow{3}{*}{Low} & \multirow{3}{*}{None}\\
         & & & & & & & &  \\
         & & & & & & & &  \\

          \hline
         \multirow{2}{*}{Stäuble et al., \cite{stauble_switz} } & \multirow{2}{*}{GIS} & \multirow{2}{*}{None} & \multirow{2}{*}{\xmark} & \multirow{2}{*}{Manual} & \multirow{2}{*}{\xmark} & \multirow{2}{*}{Medium} & \multirow{2}{*}{Low} & \multirow{2}{*}{None}\\
         & & & & & & & &  \\

          \hline
         \multirow{4}{*}{Drolias et al., \cite{inproceedings} } & \multirow{4}{*}{QGIS} & \multirow{4}{*}{Binarization} & \multirow{4}{*}{\xmark} & \multirow{4}{*}{Automated (GIS)} & \multirow{4}{*}{\checkmark} & \multirow{4}{*}{Medium} & \multirow{4}{*}{Low} & \multirow{4}{*}{Semi}\\
         & & & & & & & &  \\
         & & & & & & & &  \\
         & & & & & & & &  \\

          \hline
         \multirow{4}{*}{Gede et al., \cite{ica-proc-4-38-2021} } & \multirow{4}{1.5cm}{Open CV, Tesseract} & \multirow{4}{*}{Binarization} & \multirow{4}{*}{\xmark} & \multirow{4}{*}{Only Territory} & \multirow{4}{*}{\xmark} & \multirow{4}{*}{Medium} & \multirow{4}{*}{Low} & \multirow{4}{*}{Semi}\\
         & & & & & & & &  \\
         & & & & & & & &  \\
         & & & & & & & &  \\

         \hline
         \multirow{4}{*}{Lee et al., \cite{LEE2000165} } & \multirow{4}{1.5cm}{Open CV, GIS} & \multirow{4}{*}{Manual Edit} & \multirow{4}{*}{\xmark} & \multirow{4}{*}{Knowledge Based} & \multirow{4}{*}{\checkmark} & \multirow{4}{*}{Medium} & \multirow{4}{*}{Low} & \multirow{4}{*}{Semi}\\
         & & & & & & & &  \\
         & & & & & & & &  \\
         & & & & & & & &  \\

         \hline
         \multirow{4}{*}{Taie et al., \cite{taie2011new} } & \multirow{4}{1.5cm}{AutoCad, Commercial Software} & \multirow{4}{1.5cm}{Binarization, Noise cleaning} & \multirow{4}{*}{\xmark} & \multirow{4}{*}{Automated} & \multirow{4}{*}{\xmark} & \multirow{4}{*}{Medium} & \multirow{4}{*}{Low} & \multirow{4}{*}{Semi}\\
         & & & & & & & &  \\
         & & & & & & & &  \\
         & & & & & & & &  \\

         \hline
         \multirow{4}{*}{Iosifescu et al., \cite{Iosifescu} } & \multirow{4}{*}{GIS} & \multirow{4}{*}{Binarization} & \multirow{4}{*}{\xmark} & \multirow{4}{*}{Automated (GIS)} & \multirow{4}{*}{\checkmark} & \multirow{4}{*}{Medium} & \multirow{4}{*}{Low} & \multirow{4}{*}{Semi}\\
         & & & & & & & &  \\
         & & & & & & & &  \\
         & & & & & & & &  \\

         \hline
         \multirow{4}{*}{Ahmed et al., \cite{ahmed2024digitizing} } & \multirow{4}{*}{ML} & \multirow{4}{*}{GMM, K means} & \multirow{4}{*}{\xmark} & \multirow{4}{1.3cm}{Contrast Split Segmentation} & \multirow{4}{*}{\checkmark} & \multirow{4}{*}{Medium} & \multirow{4}{*}{Medium} & \multirow{4}{*}{Semi}\\
         & & & & & & & &  \\
         & & & & & & & &  \\
         & & & & & & & &  \\

         \hline
         \multirow{4}{*}{\textbf{Our Proposed Method} } & \multirow{4}{1.5cm}{\textbf{Open CV, GIS, ML}} & \multirow{4}{1.5cm}{\textbf{Noise Removal, Manual, Inpaint}} & \multirow{4}{*}{\checkmark} & \multirow{4}{*}{\textbf{Automated}} & \multirow{4}{*}{\textbf{\checkmark}} & \multirow{4}{*}{\textbf{Medium}} & \multirow{4}{*}{\textbf{Medium}} & \multirow{4}{*}{\textbf{Semi}}\\
         & & & & & & & &  \\
         & & & & & & & &  \\
         & & & & & & & &  \\
        
        \hline
    \end{tabular}
   \label{tab:compare_related_work}
\end{table*} 
In the following section, we address several key components of our research and provide a comparison of relevant research projects based on several criteria. We highlight and compare features such as tools used, pre-process and vectorization techniques, OCR and smoothing, computational cost, deployment effort, and automation level (none, semi or full), in \cref{tab:compare_related_work}. Most of the studies have employed GIS tools \cite{lithuanian_land, stauble_switz, inproceedings, LEE2000165, Iosifescu} while some others have used AutoCad \cite{vasile_map_digitization, east_delta_mouza}. These tools, along with OpenCV, are primarily utilized for pre-processing steps or vectorization. Binarization is the most commonly used pre-processing technique among the compared methodologies. However, some studies have used Canny Edge Detection \cite{east_delta_mouza} and unsupervised learning methods \cite{ahmed2024digitizing} such as Gaussian mixture model (GMM) and K-means clustering. During the pre-processing step, we are the first to integrate an inpaint model alongside binarization and noise removal, which is essential to reconstruct the missing parts of the rasterized mouza maps. For vectorization, unlike Piskinaite et al., \cite{lithuanian_land} and Stäuble et al., \cite{stauble_switz}, all other studies, including ours, have automated the process. Only a few of the studies have done any smoothing \cite{inproceedings, LEE2000165, Iosifescu, ahmed2024digitizing} after vectorization, which is a significant part of the automation. We have described the necessity of smoothing in \cref{subsec:vectorization}, \cref{fig:gis_output} and \cref{fig:grass_error}.

No prior research has applied Optical Character Recognition (OCR) to mouza maps, leaving a substantial portion of maps' textual information undigitized. The only exception is Taie et al., \cite{taie2011new}, who has separated the textual information from the map, however, did not digitize it. In our study, we have successfully extracted the plot numbers and employed OCR to digitize them. We also have integrated the digitized numbers with map plots (polygon), completing the digitization process. This integration of textual and spatial data is one of the key factors that distinguishes our methodology from previous works. Finally, we have evaluated the deployment effort and computational cost of the related studies. Deployment effort is estimated according to the tools used and the extent of pre-processing required. The rest of the methods mostly follow the same pattern except for those incorporating smoothing \cref{subsec:vectorization} and merging \cref{subsec:join_plot_numbers}. Both of ours and that of Ahmed et al., \cite{ahmed2024digitizing} computational cost are categorized as medium because of the inclusion of machine learning techniques. Nevertheless, our approach is computationally more efficient given the scope of tasks it accomplishes. Finally, no existing studies have achieved full automation, where a single raster image can be processed to produce a complete, smooth vector map without human intervention. Our proposed pipeline is unique in that it requires human input only during the pre-processing phase, while the remaining steps are fully automated. In contrast, previous studies require human assistance at multiple stages throughout the process. By utilizing the inpaint model, OCR, and smoothing algorithm we have achieved a semi-automated pipeline that addresses every aspect of mouza map vectorization, having humans in the loop only for the initial stage.

\subsection{Limitations and Future Work}
\label{subsec:discussion}
\begin{figure}
\centering
    \subfloat{
        \fbox{\includegraphics[width = 0.18\linewidth, height=0.17\linewidth]{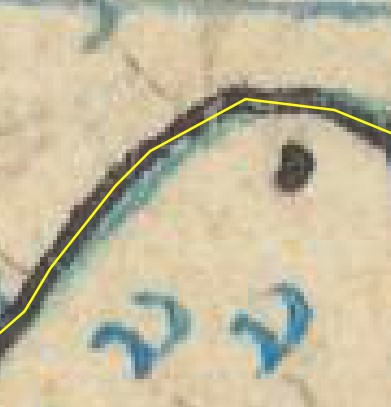}}
    }
    \hfill
    \subfloat{
        \fbox{\includegraphics[width = 0.18\linewidth, height=0.17\linewidth]{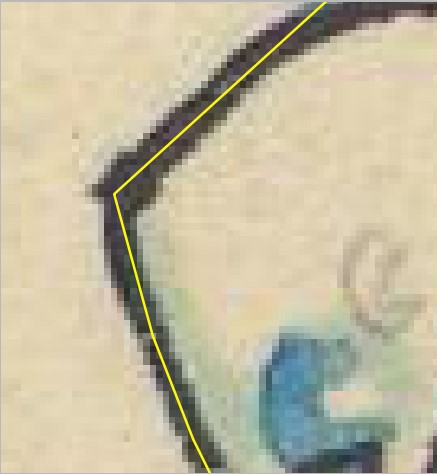}}
    }
    \hfill
    \subfloat{
        \fbox{\includegraphics[width = 0.18\linewidth, height=0.17\linewidth]{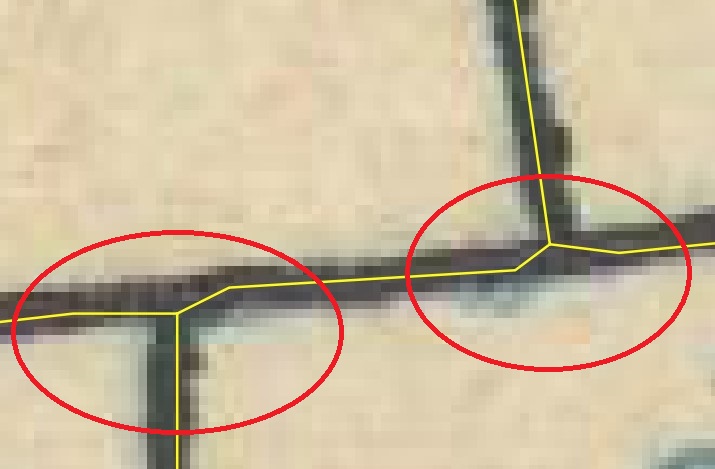}}
    }
    \hfill
    \subfloat{
        \fbox{\includegraphics[width = 0.18\linewidth, height=0.17\linewidth]{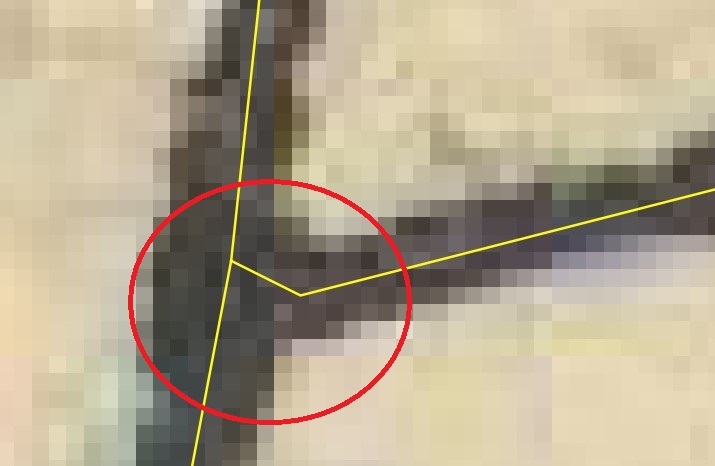}}
    }
    \\ 
    \subfloat{
        \fbox{\includegraphics[width = 0.18\linewidth, height=0.17\linewidth]{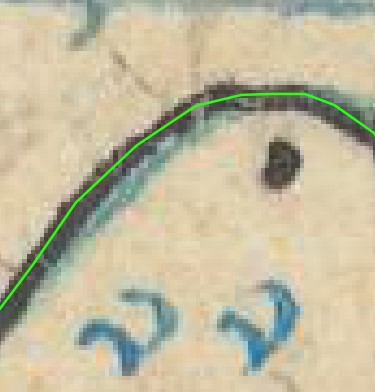}}
    }
    \hfill
    \subfloat{
        \fbox{\includegraphics[width = 0.18\linewidth, height=0.17\linewidth]{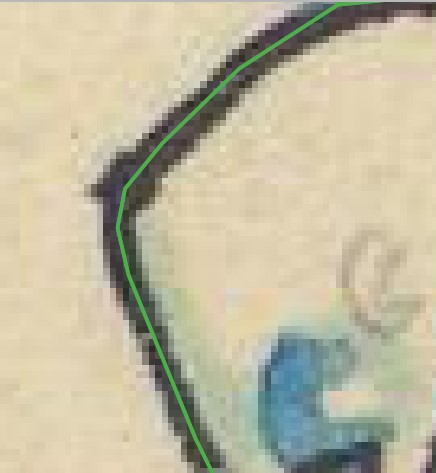}}
    }
    \hfill
    \subfloat{
        \fbox{\includegraphics[width = 0.18\linewidth, height=0.17\linewidth]{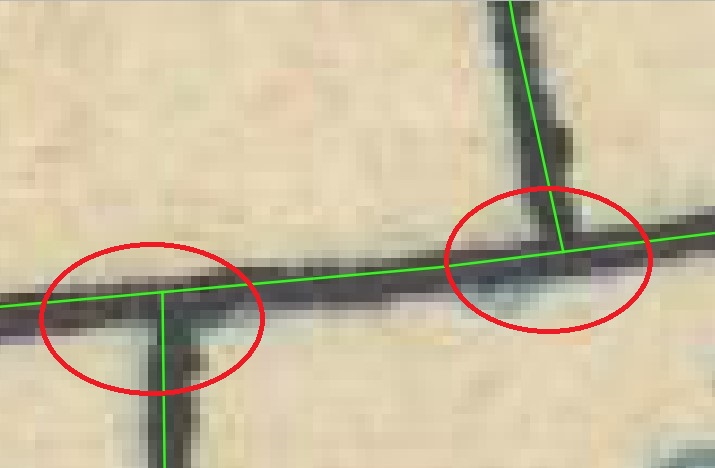}}
    }
    \hfill
    \subfloat{
        \fbox{\includegraphics[width = 0.18\linewidth, height=0.17\linewidth]{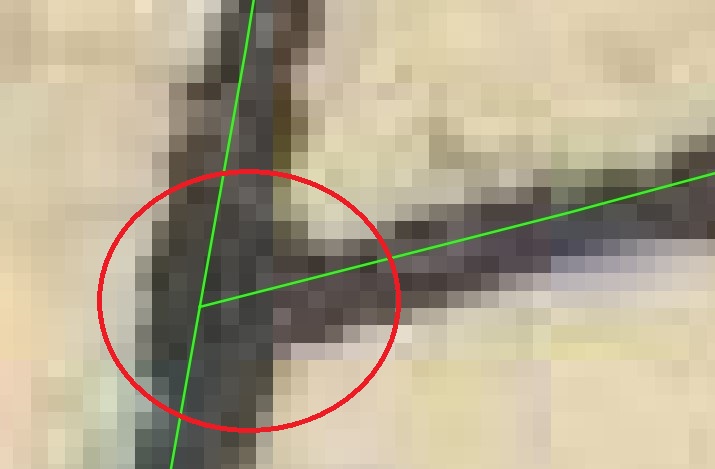}}
    }
    \caption{The first row of images are the issues we could not address with our process. The second row of images is the ground truth.}
    \vspace{-10pt}
    \label{fig:discussion}
\end{figure}
Automated vectorization of cadastral maps is a challenging task for various reasons such as complex boundary structure, quality of the source maps, high resolution, textual and spatial data integration, and accuracy requirements. To address these challenges we have proposed our methodology, which vectorizes both plot numbers and map boundary and then refine the vector output with smoothing algorithms discussed in \cref{subsubsec:smoothing}.  Despite our smoothing, certain issues persisted, notably in areas where multiple lines converge and across regions with varying curvatures. Some of these issues and the refined output after vector editing are shown in \cref{fig:discussion}. Evidently, a significant portion of the vector editing time frame was dedicated to rectifying these inaccurate vectorizations.

We employed Grass GIS \cite{GRASS_GIS_software} thinning for raster maps before vectorization and smoothing. However, this thinning process is not suitable for mouza map vectorization and causes significant distortion in the vector outputs because of the complex structure of the boundary lines. Introducing an alternative vectorization process capable of retaining contextual information from the raster map could yield more refined outputs and will substantially reduce the time required for vector editing. Another important area is pre-processing, we found that the quality of the outputs improves significantly with more thorough and careful pre-processing. To further explore, a promising avenue lies in the utilization of deep learning architectures trained on precise vector datasets of mouza maps, offering a promising advancement in this field. Moreover, the performance differences observed against the Deep Vectorizer are likely due to it being trained on a different dataset. The lack of contextual information specific to Mouza maps in its training data has led to inaccuracies in the digitization process. 

It is important to note that there is always a trade-off between time and accuracy. The comparison with the Deep Vectorizer is intended to highlight that, while the overall structure of our methodology for full digitization remains consistent, there is potential to improve our vectorization techniques, a part of our methodology. Improving these techniques could lead to better performance and significantly reduce the need for manual vector editing.
\section{Conclusion}
\label{conclusion}
In this paper, we presented a semi-automated way to digitize mouza maps while keeping humans in the loop. We discussed mouza maps and the associated challenges in their digitization. We have developed a structured methodology that systematically guides the digitization of Mouza maps through a step-by-step process. Our proposed methodology involves pre-processing, including the utilization of an inpainting model. To identify the plot numbers, separated from the boundary, we employed  Optical Character Recognition (OCR) and proceeded to vectorize the plot boundary, which is refined by our smoothing algorithms. The subsequent integration of plot numbers and vector plots completes our process. Through the implementation of these methodologies, we establish a preliminary yet significantly processed version of the vectorized map, yielding a substantial reduction in both time and effort compared to starting from scratch. In the result section we show that we achieved $15\%$ to $48\%$ time saved when editing the vector file outperforming the manual vectorization from scratch, especially with larger and more complex maps. Furthermore, we have compared our vectorization step with Deep vectorizer\cite{rassian}. Our approach completes the vectorization process in a significantly shorter time without compromising performance.

\vspace{2pt}
In conclusion, our study advances the domain of mouza map digitization through the introduction of an innovative automated methodology. This pioneering approach not only drastically diminishes the temporal and labor-intensive demands associated with manual editing but also lays the foundation for a transformative leap forward in the efficiency and precision of map digitization processes.

\vspace{2pt}
\textbf{Acknowledgment.} We express our gratitude to the Institute of Water Modeling (IWM), Bangladesh, for sharing valuable information on mouza maps and validating our outputs.

\bibliography{sample}

\vfill

\end{document}